\documentclass[10pt,twocolumn,letterpaper]{article}

\usepackage[pagenumbers]{wacv} % To force page numbers, e.g. for an arXiv version

\usepackage{graphicx}
\usepackage{amsmath}
\usepackage{amssymb}
\usepackage{booktabs}
\usepackage{pifont} %for crosses in the table
\usepackage{xcolor}
\usepackage{here}
\usepackage[super]{nth}
\usepackage{multirow}
\usepackage[page]{appendix}

\usepackage[acronym]{glossaries}
\newacronym{iou}{IoU}{Intersection over Union}
\newacronym{miou}{mIoU}{Mean Intersection over Union}
\newacronym{auc}{AUC}{Area Under the Curve}
\newacronym{got}{GOT}{Generic Object Tracking}
\newacronym{mot}{MOT}{Multiple Object Tracking}
\newacronym{gmot}{GMOT}{Generic Multi Object Tracking}
\newacronym{vos}{VOS}{Video Object Segmentation}
\newacronym{sot}{SOT}{Single Object Tracking}
\newacronym{dcf}{DCF}{Discriminative Correlation Filter}
\newacronym{fpn}{FPN}{Feature Pyramidal Network}
\newacronym{mlp}{MLP}{Multi-Layer Perceptron}
\newacronym{ope}{OPE}{One Pass Evaluation}
\newacronym{detre}{DetRe}{Detection Recall}
\newacronym{assa}{AssA}{Association Accuracy}
\newacronym{fp}{FP}{False Positive}
\newacronym{assre}{AssRe}{Association Recall}

\usepackage[pagebackref,breaklinks,colorlinks,citecolor=blue,linkcolor=blue,urlcolor=blue]{hyperref}

\newcommand{\parsection}[1]{\vspace{0.5mm}\noindent\textbf{#1.}~}

\usepackage[capitalize]{cleveref}
\crefname{section}{Sec.}{Secs.}
\Crefname{section}{Section}{Sections}
\Crefname{table}{Table}{Tables}
\crefname{table}{Tab.}{Tabs.}

\begin{document}

\title{Beyond SOT: Tracking Multiple Generic Objects at Once}

\author{Christoph Mayer$^{1}$\thanks{Work done while interning at Google Research.} \quad Martin Danelljan$^{1}$ \quad Ming-Hsuan Yang$^{2}$ \quad Vittorio Ferrari$^{2}$ \\ Luc Van Gool$^{1}$ \quad Alina Kuznetsova$^{2}$\\$^{1}$ETH Zürich \quad $^{2}$Google Research\\
\small{\textit{\{christoph.mayer, martin.danelljan, vangool\}@vision.ee.ethz.ch} \quad \textit{\{minghsuan, vittoferrari, akuznetsa\}@google.com}}}

\maketitle

%%%%%%%%% ABSTRACT
\begin{abstract}
   Generic Object Tracking (GOT) is the problem of tracking target objects, specified by bounding boxes in the first frame of a video. While the task has received much attention in the last decades, researchers have almost exclusively focused on the single object setting. However multi-object GOT poses its own challenges and is more attractive in real-world applications. We attribute the lack of research interest into this problem to the absence of suitable benchmarks.
   In this work, we introduce a new large-scale GOT benchmark, LaGOT, containing multiple annotated target objects per sequence. Our benchmark allows users to tackle key remaining challenges in GOT, aiming to increase robustness and reduce computation through joint tracking of multiple objects simultaneously. In addition, we propose a transformer-based GOT tracker baseline capable of joint processing of multiple objects through shared computation. 
   Our approach achieves a $4\times$ faster run-time in case of $10$ concurrent objects compared to tracking each object independently and outperforms existing single object trackers on our new benchmark.
   In addition, our approach achieves highly competitive results on single-object GOT datasets, setting a new state of the art on TrackingNet with a success rate \acrshort{auc} of 84.4\%. Our benchmark, code, results and trained models are available at  \mbox{\url{https://github.com/visionml/pytracking}}.
\end{abstract}

%%%%%%%%% BODY TEXT
\section{Introduction}

\begin{figure}[t]
\centering%
\includegraphics[trim = 0 0 0 0, clip, width=\columnwidth, keepaspectratio]{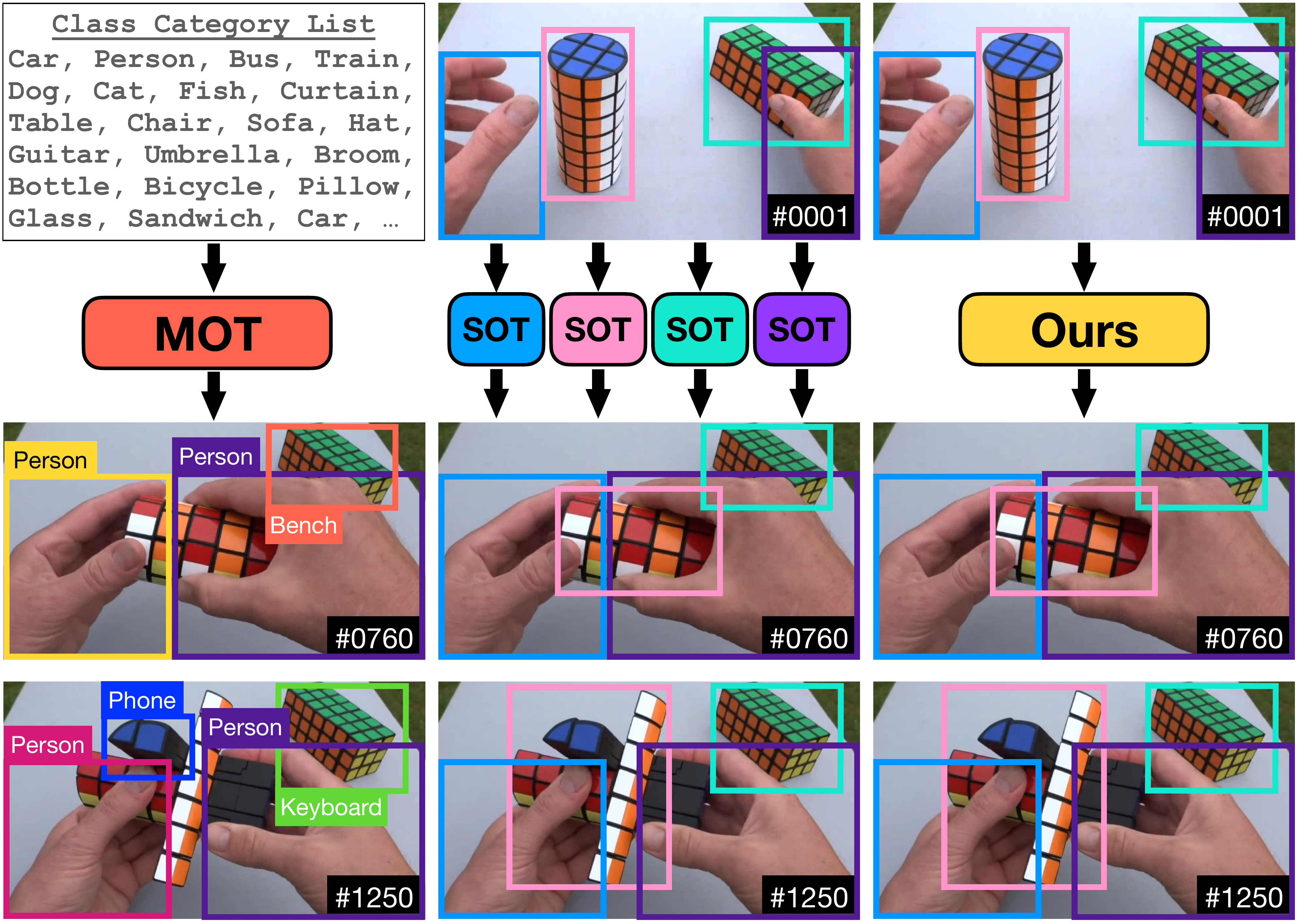}%
\vspace{-3mm}
\caption{Multiple Object trackers (MOT) track all the objects corresponding to classes in a {\em predefined} category list, while all other objects are ignored. Single Object Tracking (SOT) methods focus on tracking only a single user-specified object per video. Thus, when encountered with multiple objects, such methods must resort to independent tracking of each object. This leads to a directly linear increase in computation. Our tracker can track multiple {\em generic} objects jointly that are defined via user-specified bounding boxes, leading to the opportunity of computational savings and to exploit inter-object information for improved robustness. The box colors correspond to track IDs.}
\label{fig:teaser}
\vspace{-4mm}
\end{figure}

Visual object tracking is a fundamental problem in computer vision. Over the years the research effort has been directed mainly to two different task definitions: \gls{got}~\cite{Bertinetto_2016_ECCVW_SiameseFC,Danelljan_2017_CVPR_ECO,Henriques_2015_TPAMI_KCF,Bolme_2010_CVPR_MOSSE,Kristan_2016_TPAMI_VOT,Li_2019_CVPR_SiamRPN++,WU_2015_TPAMI_OTB} and \gls{mot}~\cite{Yu_2007_CVPR_MOT,Tang_2017_CVPR_MOT,Zhang_2008_CVPR_MOT,Dendorfer_2020_IJCV_MOTChallenge,Yu_2020_CVPR_BDD100k,Geiger_2013_IJRR_KITTI,Chang_2019_CVPR_Argoverse}.
\gls{mot} aims at detecting and tracking all objects from a predefined class category list (see Fig.~\ref{fig:teaser}), whereas all other objects are ignored.
In contrast, \gls{got} focuses on the scenario where a priori information about the object's appearance is unknown. Thus, the target model of the object's appearance must be learned at test time from a single user-specified bounding box in the initial frame, see Fig.~\ref{fig:teaser}.

While \gls{got} has a long history of active research, \gls{got} methods and benchmarks focused so far on tracking a single object per video such that the term \emph{\gls{sot}} was introduced. However, the task of \gls{got} is not limited to tracking a single object.
In fact, the ability to track multiple generic objects is desired in many real-world applications, such as surveillance, video understanding, semi-automatic video annotation, robotics, and industrial quality control.
A method that jointly tracks multiple objects can achieve substantial reduction in computational cost through shared elements, compared to running a separate instance of a \gls{sot} method for each object.
Moreover, processing multiple targets at the same time has the potential of increasing the robustness of the tracker by joint reasoning.

To facilitate the work on tracking multiple generic objects, we introduce the new multi-object \gls{got} benchmark LaGOT. It provides up to 10 user-specified generic objects in the initial frame visible through the large part of the video. The target objects in one video may correspond to completely different and previously unseen classes. Our benchmark features challenging characteristics such as fast moving objects, frequent occlusions, presence of distractors, camera motion, and camouflage. 
In total LaGOT contains 528k annotated objects of 102 different classes and an average track length of 71 seconds.

Tracking multiple target objects in the same video poses key challenges and research questions that are typically overlooked by \gls{sot} methods.
A multi-object \gls{got} method needs to jointly track multiple objects using the first-frame annotations. 
This could allow the tracker to exploit annotations of potential distractors to improve the robustness of each target model.
Furthermore, a joint localization step opens the opportunity for global reasoning across all tracks to reduce the risk of confusing similar objects.
Finally, operating on multiple local search area~\cite{Cui_2022_CVPR_Mixformer,Mayer_2022_CVPR_ToMP,Yan_2021_ICCV_STARK} is no longer feasible for a multi-object \gls{got} method  because it is inefficient and complicates re-detecting of lost objects.

We tackle these challenges by introducing a new multi object \gls{got} tracker. In order to track all desired target objects at once it operates globally by processing the full frame producing a shared feature representation for all targets. Furthermore, we propose a new generic multiple object encoding that allows us to encode multiple targets within the same training sample. We achieve this by learning a fixed size pool of different object embeddings, each representing a different target. Thus, we query the proposed model predictor with these object embeddings to produce all target models. In addition, we employ a \gls{fpn} to increase the overall tracking accuracy while operating on full-frame inputs. 

\parsection{Contributions} \textbf{(i)} We propose a novel large-scale multi-object \gls{got} evaluation benchmark, LaGOT. It provides multiple annotated objects per frame with an average of 2.9 tracks per sequence. We further evaluate several baselines on LaGOT, including two \gls{mot} and six \gls{sot} methods. We assess their quality by using \gls{got} and \gls{mot} metrics.\\
\textbf{(ii)} We develop a new baseline, TaMOs, a \gls{got} tracker that tracks multiple generic objects at the same time efficiently. To achieve this, we propose a new multi-object encoding, introduce an \acrshort{fpn} and apply the tracker globally on the entire video frame. 
TaMOs demonstrates near constant run-time when increasing the number of targets and operates at an over $4\times$ faster run-time compared to the \gls{sot} baselines when tracking $10$ objects.\\
\textbf{(iii)} We analyze TaMOs by assessing the impact of its different components using multiple benchmarks. Furthermore, TaMOs outperforms all baselines on LaGOT, while achieving excellent results on popular \gls{sot} benchmarks.
\section{Related Work}

\begin{figure*}[t]
\centering
\includegraphics[width=0.245\linewidth, keepaspectratio]{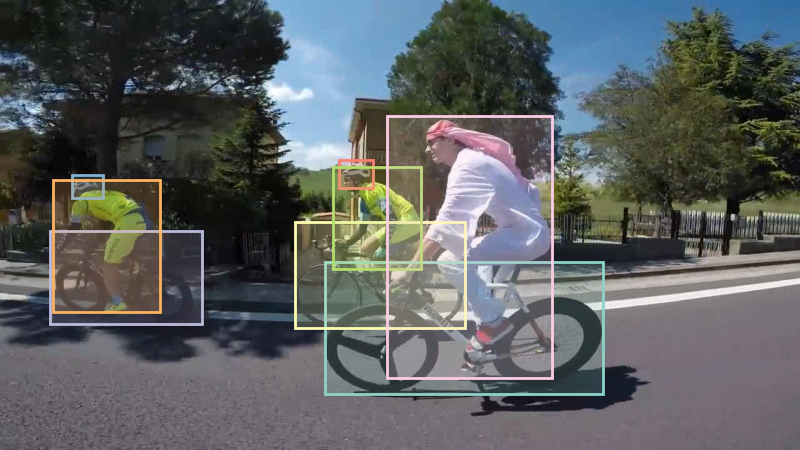}
\includegraphics[width=0.245\linewidth, keepaspectratio]{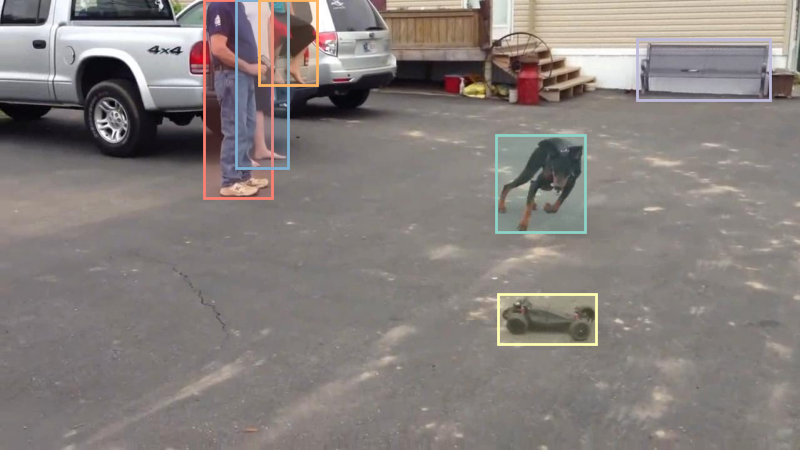}
\includegraphics[width=0.245\linewidth, keepaspectratio]{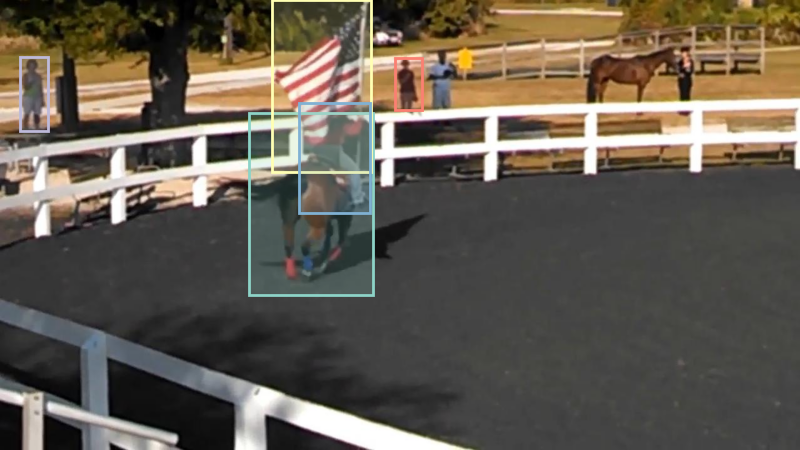}
\includegraphics[width=0.245\linewidth, keepaspectratio]{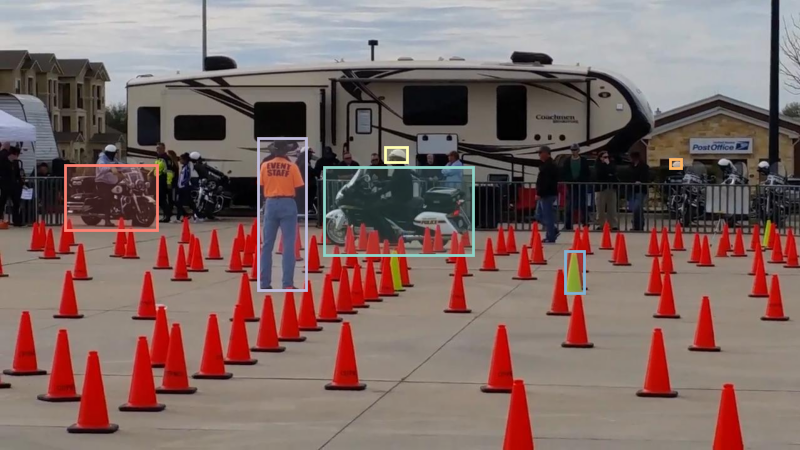}
\includegraphics[width=0.245\linewidth, keepaspectratio]{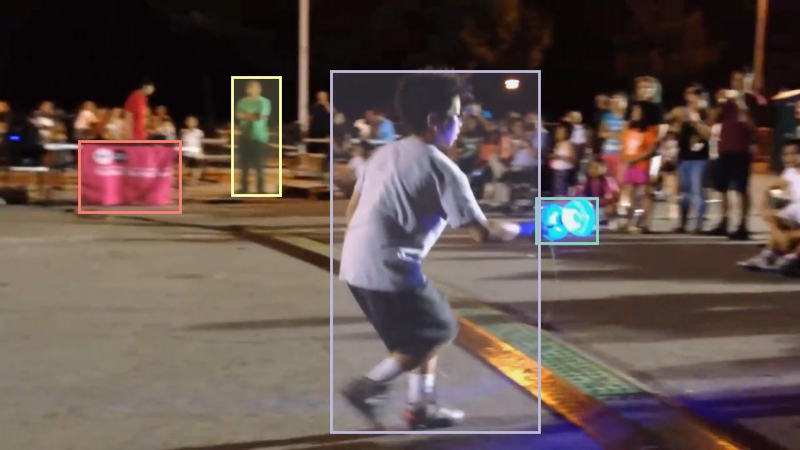}
\includegraphics[width=0.245\linewidth, keepaspectratio]{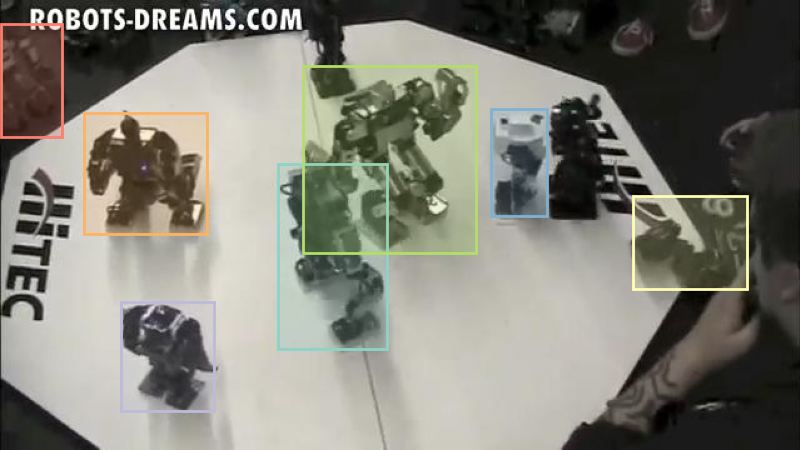}
\includegraphics[width=0.245\linewidth, keepaspectratio]{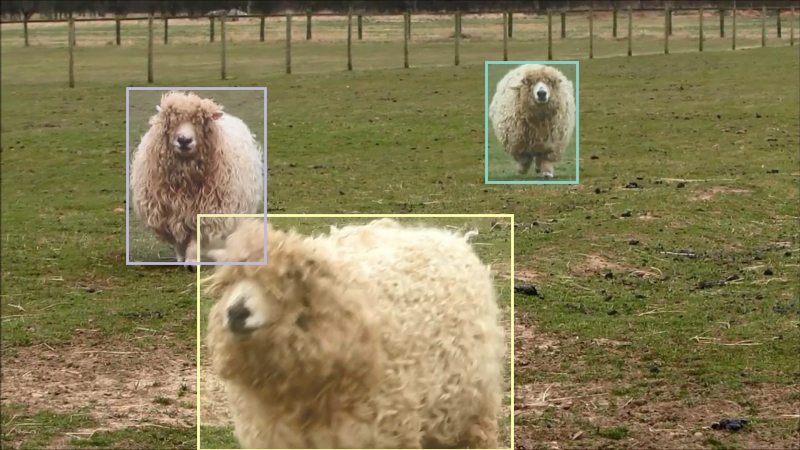}
\includegraphics[width=0.245\linewidth, keepaspectratio]{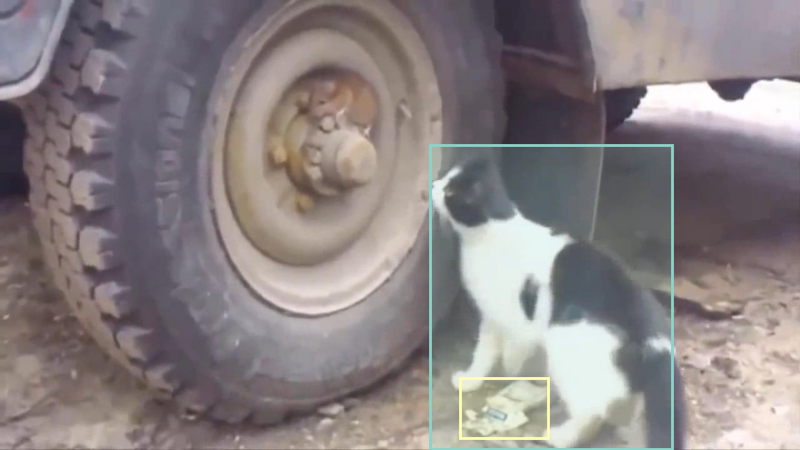}
\vspace{-3mm}
\caption{Examples of the annotated objects in the video sequences of our LaGOT dataset. The objects are annotated at $10$ FPS. Notice the diversity of the annotated media as well as the complexity of the scenes.
}\label{fig:dataset-viz}\vspace{-2mm}
\end{figure*}

\parsection{Object Tracking Benchmarks}
Generic object tracking is a well explored topic and many datasets exist. There are specialized datasets and challenges that focus on short-term~\cite{Galoogahi_2017_ICCV_NFS,WU_2015_TPAMI_OTB,Kristan_2016_TPAMI_VOT,Kristan_2020_ECCVW_VOT2020,2018_Muller_Trackingnet,Huang_2021_TPAMI_GOT10k} or long-term tracking~\cite{Fan_2019_CVPR_Lasot,Mueller_2016_ECCV_UAV123,Kristan_2020_ECCVW_VOT2020,Valmadre_2018_ECCV_OxUvA,Fan_2020_IJCV_Lasot_ext}. However, all of these benchmarks and datasets share the same setup of only providing a single user-specified bounding box such that only one target is tracked in each video sequence. Recently, GMOT-40~\cite{Bai_2021_CVPR_GMOT} focused on \gls{gmot}, where a single bounding box is provided in the first video frame and all objects that correspond to the same class as the annotated object should be tracked. In contrast to GMOT, we focus on the setting where multiple user-specified targets are given, potentially from different classes.

\gls{mot} aims at tracking multiple objects defined by a list of classes and mainly focuses on a single class~\cite{Tang_2017_CVPR_MOT,Dendorfer_2020_IJCV_MOTChallenge, huang_etal_cvpr23} (usually pedestrians) or on autonomous driving settings, where only a handful of classes are considered~\cite{Geiger_2013_IJRR_KITTI,Yu_2020_CVPR_BDD100k,Chang_2019_CVPR_Argoverse}. TAO~\cite{Achal_2020_ECCV_TAO} contains objects of a long-tailed class distributions, but provides only sparse annotations due to the costly annotation process. Another related task is open world tracking~\cite{Liu_2022_CVPR_OpenWorldTracking} that aims at detecting and tracking all objects in a video. However, compared to \gls{got} there is no mechanism to guarantee that a specific object is actually detected and tracked.In the \gls{vos} domain, DAVIS~\cite{Perazzi_2016_CVPR_Davis} and YouTubeVOS~\cite{Xu_2018_arXiv_youtubevos} provide multi-object annotations. However, their videos are extremely short ($2.9$ and $4.5$ seconds on average), and are therefore not suitable for tracking. Moreover, the \gls{vos} domain provides less challenges for trackers, instead focuses on large objects and a short-term nature, where the predominant challenge is the prediction of accurate fine-grained masks.

\parsection{Global Generic Object Tracking}
Global trackers operate on the whole video frame, rather than in a restricted search area near the object location in the previous frame.
This is not only beneficial when tracking multiple objects in the same scene but also facilitates re-detecting lost objects. GlobalTrack~\cite{Huang_2020_AAAI_GlobalTrack} and Siam R-CNN~\cite{Voigtlaender_2020_CVPR_SiamRCNN} track the target by using global RPNs that retrieve target-specific proposals. Recent method for open vocabulary tracking\cite{li2023ovtrack} tracks objects of specified classes in MOT fashion by operating on generic RPN proposals. Methods such as MetaUpdater~\cite{Dai_2020_CVPR_MetaUpdater} and SPLT~\cite{Yan_2019_ICCV_SPLT} operate on local search areas but use a re-detector to re-localize the target if it disappeared from the search area.
In contrast, our tracker TaMOs always operates on the entire frame and generates target specific correlation filters instead of target-specific proposals.

\parsection{Unified Object Tracking} Unified methods aim at tracking both: objects defined by class names or objects defined by a user-specified bounding box. UTT~\cite{Ma_2022_CVPR_UTT} allows to track all pedestrians and one generic object in each video. 
UTT uses a Transformer to match test frame features with reference features of the detected objects in the initial or previous frame.
Unicorn~\cite{Bin_2022_ECCV_Unicorn} allows to either perform the \gls{sot} or the \gls{mot} task with the same model and weights solely by varying the input data type.
In contrast, our method tracks multiple generic objects at the same time instead of one generic object or multiple objects of known classes.

\begin{table*}[t]
    \centering
    \newcommand{\dist}{\hspace{7pt}}%
    \newcommand{\yes}{\textcolor{black}{\checkmark}}
    \newcommand{\no}{\textcolor{black}{\ding{55}}}
    \caption{Comparison of LaGOT with existing benchmarks that focus on related tasks to multi-object GOT.}
    \vspace{-3mm}
    \label{tab:dataset-comp}%
    \resizebox{0.99\linewidth}{!}{%
        \begin{tabular}{l@{\dist}|c@{\dist}c@{\dist}c@{\dist}c@{\dist}|c@{\dist}c@{\dist}c@{\dist}c@{\dist}c@{\dist}c@{\dist}c@{\dist}}
        \toprule
                                                 &      & Object     & Num Classes & Tracking    & Num     & Num    & Avg Video & Avg Track & Avg Tracks & Num         & Annotation \\
         Dataset                                 & Task & Definition & per Video   & Metrics   & Classes & Videos & Length (s)    & Length (num anno.)   & per Video  & Annotations & Frequency  \\
        \midrule
        TAO val~\cite{Achal_2020_ECCV_TAO}       & MOT  & class list & $\geq 1$    & Track-mAP   & 302     & 988    & 33.5      & 21       & 5.55       & 115k        & 1 FPS      \\ 
        \midrule
        GMOT-40~\cite{Bai_2021_CVPR_GMOT}        & GMOT & 1 box      & 1           & MOTA/IDF1   & 30      & 40     & 10       & 133       & 50.65      & 486k        & 24-30 FPS  \\ 
        \midrule
        LaSOT val~\cite{Fan_2019_CVPR_Lasot}     & SOT  & 1 box      & 1           & Success AUC & 70      & 280    & 81      & 2430      & 1          & 680k        & 30 FPS     \\
        \midrule
        \textbf{LaGOT}                           & GOT  & n boxes    & $\geq 1$    & F1-Score    &102     & 294    & 75.3      & 707      & 2.89       & 528k        & 10 FPS     \\
        \bottomrule
        \end{tabular}
	}\vspace{-4mm}
\end{table*}

\parsection{Transformers for Generic Object Tracking}
Tracking has seen a tremendous progress in recent years with the advent of Transformers~\cite{Vaswani_2017_NIPS_ATTENTION}.
Most such trackers share the idea of fusing the search area and the template image features by using a Transformer~\cite{Cui_2022_CVPR_Mixformer,Yan_2021_ICCV_STARK,Chen_2021_CVPR_TransT,Mayer_2022_CVPR_ToMP,Ye_2022_ECCV_OSTrack,Yu_2021_ICCV_HPF}. MixFormer~\cite{Cui_2022_CVPR_Mixformer} and OSTrack~\cite{Ye_2022_ECCV_OSTrack} employ a Transformer to jointly extract and fuse the template and search area features.
TransT~\cite{Chen_2021_CVPR_TransT}, STARK~\cite{Yan_2021_ICCV_STARK}, SwinTrack~\cite{Lin_2021_NeurIPS_SwinTrack} and ToMP~\cite{Mayer_2022_CVPR_ToMP} use a backbone to extract features and employ cross attention to fuse the feature representations.
However, none of these trackers can easily be extended to jointly track multiple objects, which is addressed in this work. 
\section{LaGOT Benchmark}

In this section we first introduce the multi-object \gls{got} task and discuss its differences to other object tracking tasks. Then, we introduce our new benchmark LaGOT.

\subsection{Multi-object GOT Task}

Multi-object \gls{got} is the task of tracking multiple generic target objects in a video sequence. The target objects are defined by user-specified bounding boxes in the initial frame of the video. Thus, the target objects are generic in the sense that their class category is unknown and there might be no object of the same category in the training data, see Fig.~\ref{fig:teaser}.

\parsection{Multi-object GOT vs.\ SOT}
\gls{sot} requires to track only a single target object defined by the user~\cite{WU_2015_TPAMI_OTB,Fan_2019_CVPR_Lasot,Kristan_2016_TPAMI_VOT}, whereas multi-object \gls{got} focuses on tracking multiple user-specified generic target objects in the same video.

\parsection{Multi-object GOT vs.\ MOT}
\textbf{(i)} The \gls{mot} task requires to track all objects of known classes, whereas for multi-object \gls{got} target objects in each video are defined by user-specified boxes. Consequently, multi-object \gls{got} is a one-shot problem where the target objects are unknown at training time and are only available during inference. In contrast, traditionally \gls{mot} methods track all objects corresponding to the categories defined at training time.
\textbf{(ii)} For the multi-object \gls{got} task an object-id switch is equivalent to a complete failure since the user-specified object is no longer recoverable~\cite{WU_2015_TPAMI_OTB,Lukezic_2018_arXiv_VOTLT_Metric}. Conversely, for \gls{mot} methods object-id switches are considered less problematic and are penalized less drastically by the \gls{mot} metrics~\cite{Luiten_2021_IJCV_HOTA}.

\parsection{Multi-object GOT vs.\ GMOT}
\gls{gmot} focuses on tracking multiple objects of a single generic object class in each video. The class is defined by a single user-specified bounding box in the initial video frame~\cite{Ferryman_2009_IWPETS_PETS,Bai_2021_CVPR_GMOT}. Thus, in contrast to multi-object \gls{got}, a \gls{gmot} method is unable to track multiple objects of different categories in the same video.

\subsection{LaGOT}

\parsection{Benchmark Construction}
LaSOT~\cite{Fan_2019_CVPR_Lasot} contains diverse and relatively long videos (2430 frames or 81 seconds on average) with challenging tracking scenarios including fast moving objects, camera motion, various object sizes, frequent object occlusions, scale changes, motion blur, camouflage and objects that go out of view or change their appearance. LaSOT provides annotations for a single object in each video but typically multiple objects are present throughout the full sequence and are fairly difficult to track, which is desirable for long-term tracking scenarios.
Thus, instead of collecting new videos, we used the popular LaSOT evaluation set and add new annotations for multiple objects in each sequence.

Another large-scale video dataset we considered is TAO~\cite{Achal_2020_ECCV_TAO} and GMOT-40~\cite{Bai_2021_CVPR_GMOT}. However, compared to LaSOT, TAO contains shorter videos with an average of 33 seconds and its outdoor and road sequences mainly focus on pedestrians and vehicles (60\% of all objects in TAO). While the indoor sequences contain rarer object categories, they are often static and are only visible for a short time. Furthermore, TAO contains only sparse annotations (1 FPS). For all these reasons, we used LaSOT instead of TAO to build our benchmark.
GMOT-40~\cite{Bai_2021_CVPR_GMOT} contains dense annotations, but videos often contain many objects of a single class. Furthermore, GMOT-40 consists of only 40 short sequences (avg 240 frames or 8 seconds) rendering only 10 different object classes, see Tab.~\ref{tab:dataset-comp}. Thus, GMOT-40 is unsuitable to serve as a multi-object \gls{got} benchmark. 

\begin{figure*}[t]
\centering
\includegraphics[width=0.85\textwidth, keepaspectratio]{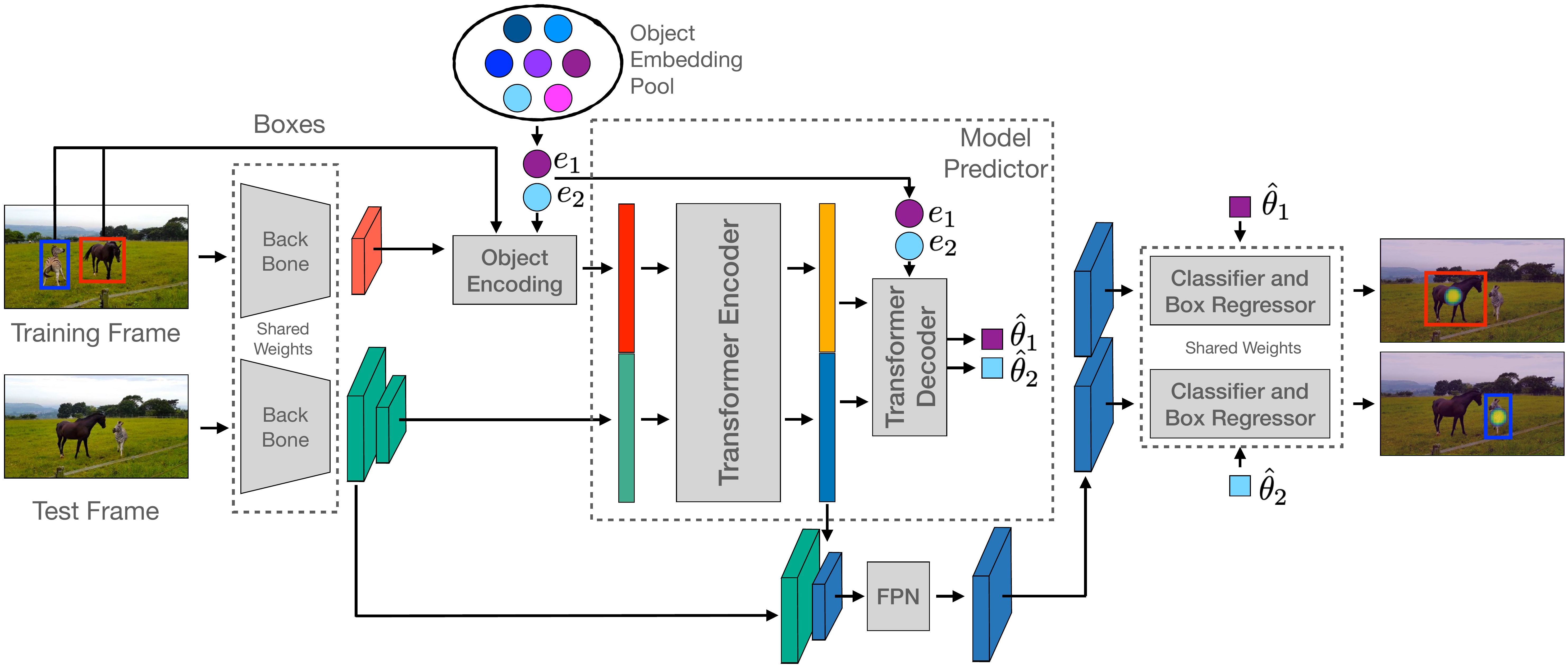}
\vspace{-3mm}
\caption{Overview of our tracker TaMOs for joint tracking of multiple targets. First, we extract features from training and test frames. All objects in the training frame are encoded jointly with a multi-object encoding and passed to the model predictor together with the training frame features. The model predictor produces target models $\hat{\theta_i}$ together with enhanced test features. We apply an \acrshort{fpn} on the enhanced output features to generate higher resolution test features. Finally, we predict the bounding box of each target by applying the target model $\hat{\theta_i}$ for each target.
}\label{fig:diagram}
\vspace{-5mm}
\end{figure*}

\parsection{Annotation Protocol}
First, we inspect all $280$ sequences in LaSOT and identify in each video challenging target objects that play an active role and meet the previously specified criteria. Next, we entrust professional annotators to annotate the selected objects in all sequences on every third frame, leading to an annotation frequency of $10$ FPS.
They use an interactive annotation tool which incorporates an object tracker to speed up the annotation process~\cite{Kuznetsova_2021_WACV}. A group of researchers verifies the newly obtained annotations and sends low-quality annotations back for correction until all annotations meet our high quality standards.
Finally, we post-process the annotations to construct the final tracks. First, we remove all tracks shorter than $4$ seconds. Second, we define the starting frame by manually selecting the earliest frame where as many annotated objects as possible are clearly visible. Third, it is not always possible to unambiguously associate all object identities over time due to occlusions and out-of-view events --- hence, we either remove ambiguous annotations or cut these videos into multiple sub-sequences, where the object association is clear. We follow this protocol to guarantee a high annotation quality, see Fig.~\ref{fig:dataset-viz} for annotated example frames.

\parsection{Statistics} Our benchmark LaGOT has 294 videos with 850 tracks leading to over 528k annotated objects. Thus, we almost triple the number of tracks compared to the original LaSOT validation set (and the corresponding evaluation time from 378 to 1006 min).
Furthermore, we add 31 additional generic object classes, \eg propeller, tires or fabric bag. 
We compare the proposed benchmark with the most closely related benchmarks in Tab.~\ref{tab:dataset-comp}~(and with many more Tab.~2 in suppl. material).
Overall our benchmark contains $10\times$ more class categories than GMOT-40.
The average track length of LaGOT is $2121$ frames ($707$ annotated frames), which is $3\times$ longer than in TAO, and  almost $10\times$ longer than in GMOT-40. 

\parsection{Annotation Frequency}
According to Valmadre~\etal~\cite{Valmadre_2018_ECCV_OxUvA} it is more effective to spend a fixed annotation budget on many videos with sparse box annotations than on fewer videos with dense labels. Thus, we annotate every third frame to reduce the overall annotation cost. To analyze the difference between 10 and 30 FPS annotations, we evaluate five recent trackers on the tracks borrowed from LaSOT, where 30 FPS annotations are available. The mean relative error of the success rate AUC is only 0.237\%. This shows that 10 FPS is sufficient on large-scale datasets such as LaSOT and LaGOT, leading to only minor score deviations.
\section{Method}

In this section we present our tracker TaMOs, which employs a Transformer to jointly model and track a set of arbitrary objects defined in the initial frame of a video.
We start from ToMP~\cite{Mayer_2022_CVPR_ToMP}, a recent Transformer-based generic single object tracker that operates on local search area cropped from the full frame, as almost all \gls{sot} trackers.
ToMP employs a transformer to predict a correlation filter (target model) from the target appearance in the initial frame conditioned on the new frame; the predicted target models is later used to localize the target in the subsequent frames. In Sec.~\ref{sec:overview} we introduce the proposed Transformer-based multi-object tracking architecture and in Sec.~\ref{sec:training-inference} we discuss the used training protocol. 

\subsection{Generic Multi-Object Tracker - Overview}\label{sec:overview}
An overview of the proposed generic multi-object tracker TaMOs is presented in Fig.~\ref{fig:diagram}.
First, unlike original ToMP, our tracker operates on the full train and test images instead of crops. The target object encoder uses a pool of learnable object embeddings to encode the location and extent of each target object within a single shared feature map (Sec.~\ref{sec:encoding}). The randomly sampled object embedding then represents a particular target in the entire video sequence: we use the object embedding to condition the model predictor to produce the target model that localizes the target object in the test frame (Sec.~\ref{sec:model-prediction}). 
Since operating on the entire video frame increases the computational cost of the Transformer operations, we are limited to a certain feature resolution. 
To track small objects we propose an \acrshort{fpn}-based feature fusion of the test frame features produced by the Transformer with the higher resolution backbone features.
We adopt the correlation filter based target localization and bounding box regression mechanism of ToMP but apply both on the higher resolution \gls{fpn} features instead of the output features of the Transformer (Sec.~\ref{sec:loc_and_fpn}).

\subsubsection{Generic Multiple Object Encoding}\label{sec:encoding}
To track several target objects efficiently, we propose a novel object encoding to embed multiple objects in a shared feature map without requiring multiple templates. 
In particular, we extend the single object encoding formulation of ToMP to be applicable for multiple objects. The idea is to replace the foreground embedding with multiple object embeddings, each representing a different target object.
Thus, we create a pool $E\in\mathbb{R}^{m \times c}$ of $m$ object embeddings $e_i \in \mathbb{R}^{1\times c}$. Then, we sample for each target object a random object embedding from the pool $E$ without replacement.
Next, we combine the object embeddings with the Gaussian score map $y_i \in \mathbb{R}^{h\times w \times 1}$ that represents the center location of the target object $i$ and the LTRB~\cite{Tian_2019_ICCV_FCOS,Xu_2020_AAAI_SiamFCpp} bounding box encoding $b_i^{\mathrm{ltrb}} \in \mathbb{R}^{h\times w \times 4}$. The final encoding is thus:
\begin{equation}\label{eq:mo_encoding}
    f_\mathrm{train}^\mathrm{enc} = f_\mathrm{train} + \sum_{i = 0}^{n} e_i \cdot y_i + \sum_{i=0}^{n} e_i \cdot \phi\left(b_i^{\mathrm{ltrb}}\right),
\end{equation}
where $f_\mathrm{train} \in \mathbb{R}^{h\times w \times c}$ are visual features extracted from the full training frame, $\phi$ is a \gls{mlp} and $n \leq m$ is the number of tracked objects. Note, that in contrast to the object encoding in ToMP, we not only use the object embedding to encode the Gaussian score map but also the bounding box representation. The object embeddings $e_i$ are learned during training such that the model is able to disentangle the shared feature representation and can identify each object in the training and test features.
Note, that the products in Eq.~\eqref{eq:mo_encoding} employ multiplications with broadcasting across every dimension whereas the latter uses channel-wise multiplication with broadcasting across the spatial dimensions.

\subsubsection{Joint Model Prediction}\label{sec:model-prediction} 

Now that the target object locations and extents are embedded in the training features, we require a model predictor to produce a target model for each encoded object. The target models are then used to localize the targets in the test frame and to regress their bounding boxes. In order to easily associate the different targets over time, we require a model predictor that can be conditioned on the targets encoded through object embeddings $e_i$. Furthermore, the model needs to be able to produce all target models jointly to increase the efficiency.

We extend the single target model predictor of ToMP by keeping the Transformer encoder unchanged but by modifying the Transformer decoder. In particular, we query the Transformer decoder with multiple object embeddings $e_i$ at the same time instead of a single foreground embedding,
\begin{equation}
    [\hat{\theta}_1, \dots \hat{\theta}_n] = T_\mathrm{dec}([h_\mathrm{train}, h_\mathrm{test}], [e_1, \dots e_n]) \,.
\end{equation}
Here, $\hat{\theta_i}\in \mathbb{R}^{c}$ is the target model, $n$ is the number of target objects encoded in the training frame and $h_\mathrm{train}$, $h_\mathrm{test}$ are the refined output features of the Transformer encoder for the train and test frame.

\subsubsection{Target Localization and Box Regression}\label{sec:loc_and_fpn}

We use the generated target models to localize the targets and to regress their bounding boxes. We produce a correlation filter for target classification and adopt the bounding box regression branch of ToMP~\cite{Mayer_2022_CVPR_ToMP}. But instead of applying the target classifier and box regressor on the low-resolution test features $h_\mathrm{test}$ of the Transformer encoder, we use high resolution features generated with an \gls{fpn} $\psi(\cdot)$ and obtain the high-resolution multi-channel score map:
\begin{equation}
    \hat{y}^\mathrm{high}_i = w_i^\mathrm{cls}(\hat{\theta}_i) \ast \psi(h_\mathrm{test}, f^\mathrm{high}_\mathrm{test}), \quad 0\leq i < n,
\end{equation}
where $f^\mathrm{high}_\mathrm{test}\in\mathbb{R}^{2h\times 2w \times c}$ are the high-resolution test features extracted at an earlier stage of the backbone, $w_i^\mathrm{cls}(\hat{\theta}_i)$ refers to the discriminative correlation filter for the target object $i$ obtained from the predicted target model $\hat{\theta}_i$. Similarly we obtain the high-resolution multi-channel bounding box regression maps $\hat{b}_i^\mathrm{high}$.

\subsection{Training}
\label{sec:training-inference}

During training we employ a classification and a bounding box regression loss.
We compute both losses for the predictions obtained by processing each \gls{fpn} feature map (low-res and high-res) as well as the output test features $h_\mathrm{test}$ of the Transformer encoder. 
The classification loss is
\begin{equation}
    L_\mathrm{cls} = \sum_{i=0}^{n}L_\mathrm{focal}(\hat{y}_i, y) + \sum_{j=n}^{m}L_\mathrm{focal}(\hat{y}_j, 0),
    \label{eq:loss}
\end{equation}
Here we assume that the first $n$ object embeddings $e_i$ were used to encode the $n$ objects marked in the training frame whereas the remaining $m-n$ object embeddings were not used to encode any objects. Thus, we require that the resulting score maps $\hat{y}_j$ that correspond to an unused object embedding $e_j$ produce low scores everywhere (second sum in Eq.~\eqref{eq:loss}). This step tightly couples the object encoding and decoding. Omitting this term not only decreases the overall performance but slows down the training progress.

In contrast to classification, we enforce the generalized IoU-Loss~\cite{Rezatofighi_2019_CVPR_GIOU} for bounding box regression only for the predictions that actually correspond to an encoded object and ignore those corresponding to unused object embeddings. 

\parsection{Training Details}
We randomly sample an image pair consisting of one training and one test frame from a training video.
The frames are re-scaled and padded to a resolution of $384\times576$. 
We train our tracker on the training splits of LaSOT~\cite{Fan_2019_CVPR_Lasot}, GOT10k~\cite{Huang_2021_TPAMI_GOT10k}, TrackingNet~\cite{2018_Muller_Trackingnet},  MS-COCO~\cite{Lin_2014_ECCV_COCO}, ImageNet-Vid~\cite{Russakovsky_2015_IJCV_ILSVRC15}, TAO~\cite{Achal_2020_ECCV_TAO}, and YoutubeVOS~\cite{Xu_2018_arXiv_youtubevos}. Note, that we remove all videos from the TAO training set that overlap with the evaluation set of LaSOT.
We randomly sample for each epoch 40k image pairs with equal probability from all datasets.
In order to leverage SOT datasets and training all object embeddings $e_i$ equally, we assign random object ids to all objects in the sampled training pair.
Note, that both SOT and MOT datasets are crucial to train the proposed tracker. Without MOT datasets the tracker is unable to learn multiple target models at the same time and avoiding SOT datasets leads to inferior tracking quality. 
We train the tracker for 300 epochs on 4 Nvidia A100 GPUs. Our method is implemented using PyTracking~\cite{Danelljan_2019_github_pytracking}
(see suppl.\ material for further details).
\section{Experiments}\label{sec:evaluation}
\begin{figure*}[t]
\centering
\begin{subfigure}{0.49\linewidth}
    \includegraphics[width=\columnwidth, keepaspectratio]{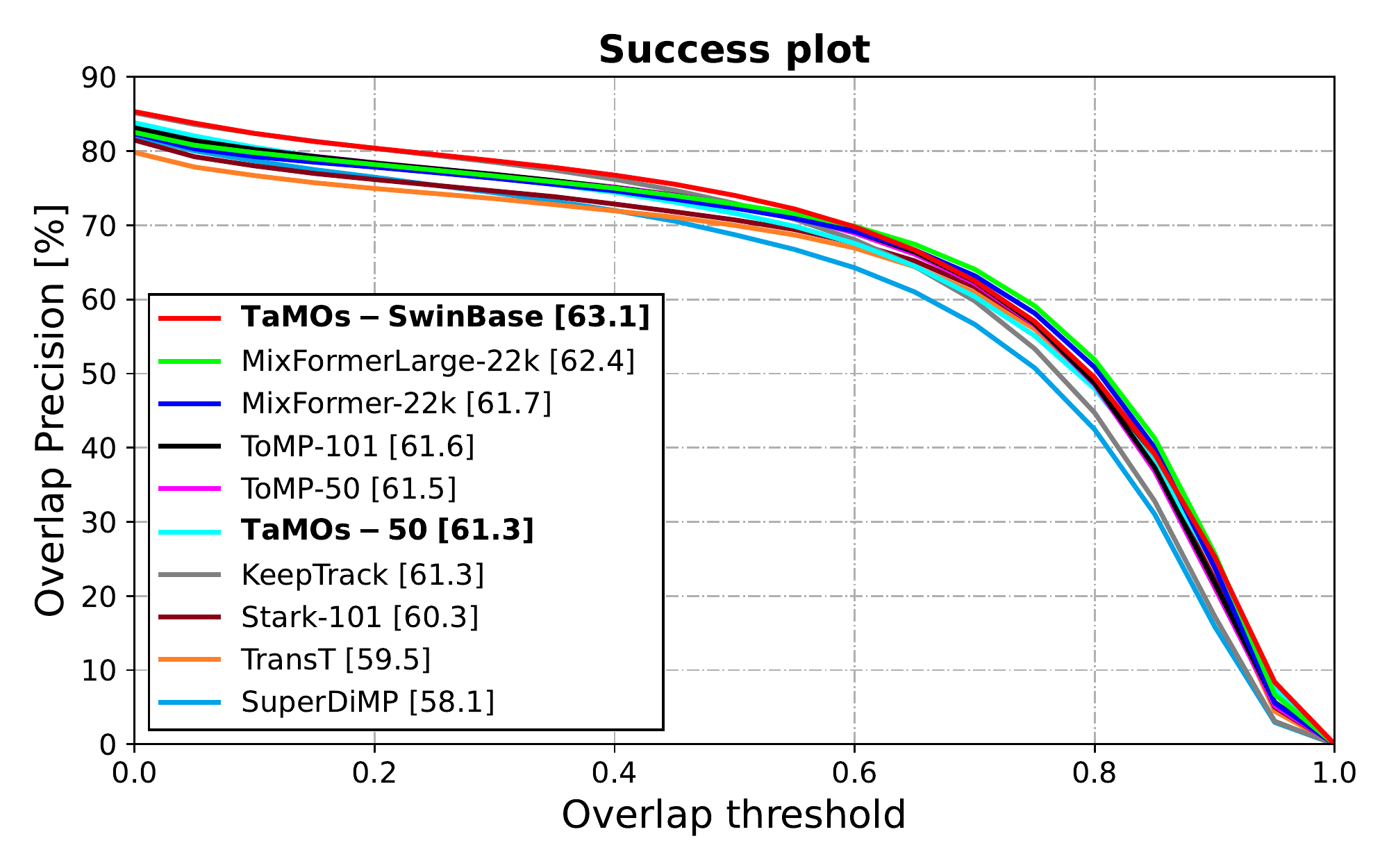}\vspace{-2mm}
    \caption{Success Plot\vspace{-2mm}}
    \label{fig:success}
  \end{subfigure}
  \hfill
  \begin{subfigure}{0.49\linewidth}
    \includegraphics[width=\columnwidth, keepaspectratio]{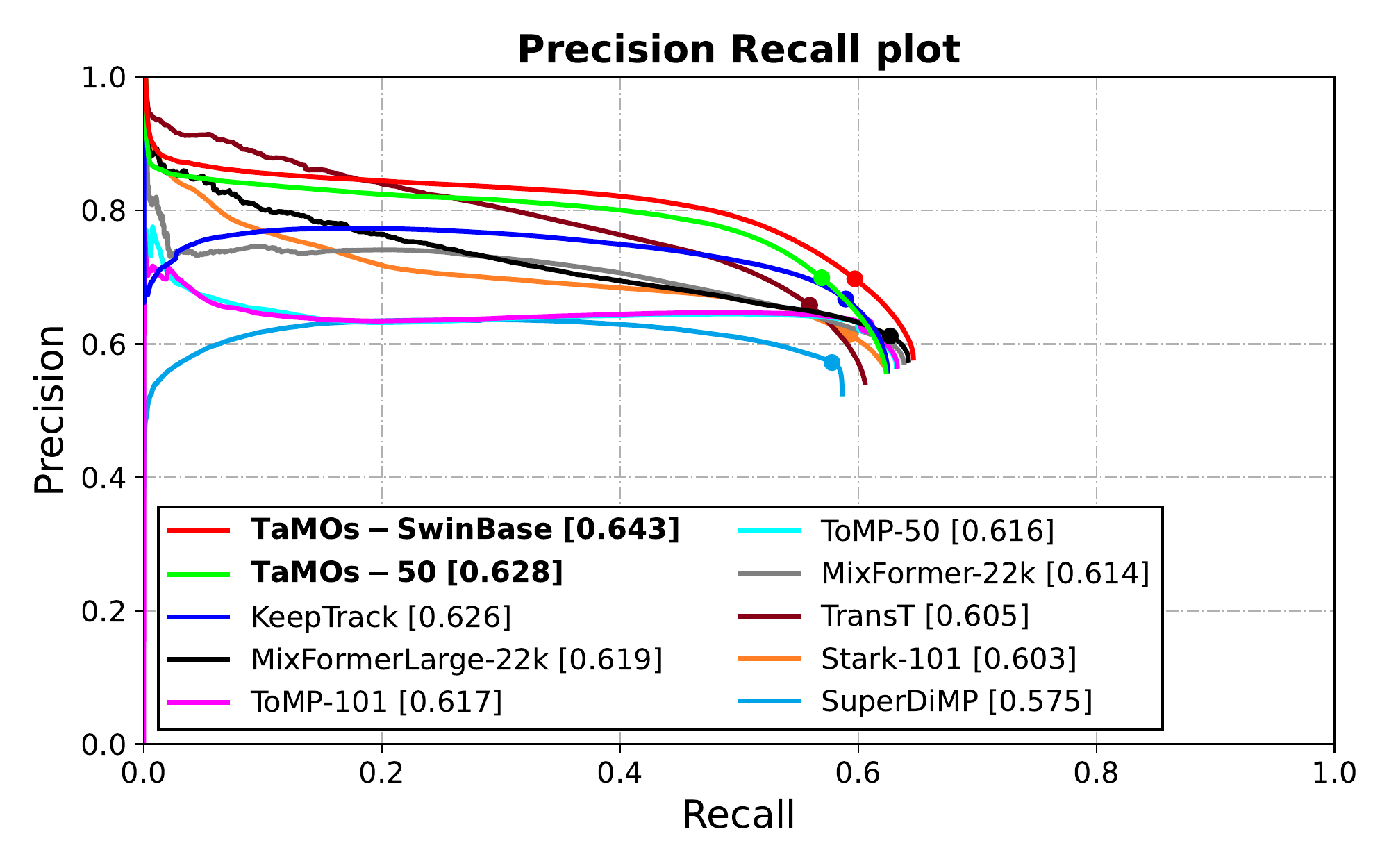}\vspace{-2mm}
    \caption{Precision-Recall Plot\vspace{-2mm}}
    \label{fig:prec_rec}
  \end{subfigure}
\caption{Success plot, showing $\text{OP}_T$, on LaGOT (AUC is reported in the legend). Tracking Precision-Recall curve on LaGOT -- VOTLT  is reported in the legend (the highest F1-score).}\label{fig:lagot_results}\vspace{-4mm}
\end{figure*}

To illustrate the challenges of our proposed \gls{got} benchmark,
we evaluate several recent trackers along with our proposed tracker TaMOs on LaGOT (Sec.~\ref{subsec:eval_lagot}). In addition, we compare TaMOs to recent trackers on several \gls{sot} benchmarks (Sec.~\ref{subsec:sot_experiments}) and present an ablation study (Sec.~\ref{subsec:ablation}), evaluating the impact of different components of our tracker.

\subsection{State-of-the-Art Evaluation on LaGOT}\label{subsec:eval_lagot}

We evaluate our tracker with a ResNet-50 and a SwinBase backbone as well as six single object trackers (SuperDiMP~\cite{Danelljan_2019_github_pytracking}, KeepTrack~\cite{Mayer_2021_ICCV_KeepTrack}, TransT~\cite{Chen_2021_CVPR_TransT}, STARK~\cite{Yan_2021_ICCV_STARK}, ToMP~\cite{Mayer_2022_CVPR_ToMP}, and MixFormer~\cite{Cui_2022_CVPR_Mixformer}) and two multi object trackers (QDTrack~\cite{Pang_2021_CVPR_QDTrack} and OVTrack~\cite{li2023ovtrack}) on LaGOT.

\parsection{Metrics} 
We measure the performance of a tracker in the \gls{ope} setting.
The standard \gls{got} Success rate \gls{auc} metric~\cite{WU_2015_TPAMI_OTB,Fan_2019_CVPR_Lasot,Fan_2020_IJCV_Lasot_ext,2018_Muller_Trackingnet,Galoogahi_2017_ICCV_NFS,Noman_2022_BMVC_AVisT,Mueller_2016_ECCV_UAV123}.
does not account for false positive predictions when a target gets occluded or is out of view. While this is not a big issue in standard \gls{sot} datasets, where the target object is present in the vast majority of frames, it becomes vital in long-term tracking.
In LaGOT objects are more frequently invisible due to occlusions or moving out-of-view.
To capture this aspect, we employ the VOTLT~\cite{Matej_2018_ECCVW_VOT2018,Lukezic_2018_arXiv_VOTLT_Metric} metric that penalizes false positives. It computes the IoU-weighted precision-recall curve and ranks the trackers according to their F1-score. 
\vspace{-0.2cm}
\subsubsection{Comparison to SOT Methods}
\gls{sot} trackers are limited to track only a single target at once. Thus, multiple instances of the same tracker need to be run in parallel to track multiple objects in the same sequence leading to a linearly increasing run-time, see Fig.~\ref{fig:teaser}. 
\parsection{Results}
Fig.~\ref{fig:success} shows the success rate of all trackers on LaGOT. We observe that \gls{sot} trackers perform well on LaGOT. However, our multi-object tracker TaMOs achieves the best \gls{auc}, even outperforming the state-of-the-art \gls{sot} tracker MixFormerLarge-22k~\cite{Cui_2022_CVPR_Mixformer}.
We further observe that TaMOs is as robust as KeepTrack~\cite{Mayer_2021_ICCV_KeepTrack} ($T < 0.4$), where the gap to the remaining trackers is particularly prominent.
This demonstrates the potential of a global multiple object \gls{got} method.
Fig.~\ref{fig:prec_rec} shows the \textit{tracking} Precision-Recall curve on LaGOT. Both versions of TaMOs outperform all other \gls{sot} trackers. The highly robust object presence scores predicted by our tracker lead to a superior precision at all recall rates $>0.2$. Moreover, our approach achieves the best maximal recall and outperforms all previous methods in VOTLT by $1.7$ points.
This demonstrates that joint tracking of multiple objects and global search benefit the object localization and identification capabilities of the tracker.
For further insights we show \gls{mot} metrics on LaGOT in Tab.~\ref{tab:metrics_lagot}. Our tracker achieves the best results for every \gls{mot} metric and outperforms MixFormerLarge-22k by $5.9$ points in MOTA.
\parsection{Run-Time Analysis}
We evaluate the run-time on a single A100 GPU.
Tab.~\ref{tab:run_time} reports a run-time analysis of our tracker TaMOs compared to ToMP, with both employing a ResNet-50 backbone.
While TaMOs is slower than ToMP for a single object, due to the higher resolution required for full-frame tracking, our approach already reaches an advantage for 2 concurrent objects. As ToMP needs to run a separate independent tracker for each new object, our approach achieves a $4\times$ speedup for 10 concurrent objects. Furthermore, the analysis demonstrates that TaMOs achieves almost a constant run-time even when increasing the number of targets.
TaMOs-SwinBase achieves 13.1 FPS for a single object and 9.3 FPS when jointly tracking $10$ objects.
\vspace{-0.2cm}
\subsubsection{Comparison to MOT Methods}
\gls{mot} methods are designed to track multiple objects in a video sequence and are thus used as baselines for LaGOT. 
However, in \gls{mot} the targets are defined via a list of classes whereas in multi-object \gls{got} targets are defined by user-specified bounding boxes in the initial frame.
Hence, to be able to track generic objects \gls{mot} methods need to be trained on large vocabulary datasets --- then we can greedily match the detected tracks with the bounding boxes on the initial frame to track user-specified objects.
Alternatively, the recent open-vocabulary MOT
method OVTrack~\cite{li2023ovtrack} allows to track objects of any class. We employ QDTrack~\cite{Pang_2021_CVPR_QDTrack} and open-vocabulary OVTrack~\cite{li2023ovtrack} as \gls{mot} baselines. QDTrack is trained on LVIS~\cite{Gupta_2019_CVPR_LVIS} and TAO. We provide OVTrack in each video with the class name of the target.
\begin{table}[!t]
    \centering
    \newcommand{\best}[1]{\textbf{\textcolor{red}{#1}}}
    \newcommand{\scnd}[1]{\textbf{\textcolor{blue}{#1}}}
    \newcommand{\dist}{\hspace{20pt}}%
    \newcommand{\yes}{\textcolor{black}{\checkmark}}
    \newcommand{\no}{\textcolor{black}{\ding{55}}}
    \caption{Run-time analysis (in FPS) between our baseline model ToMP and our tracker TaMOs.}%
    \vspace{-3mm}
    \resizebox{\columnwidth}{!}{%
        \begin{tabular}{l@{\dist}c@{\dist}c@{\dist}c@{\dist}c@{\dist}}
            \toprule
                            & 1 Object & 2 Objects & 5 Objects & 10 Objects\\
            \midrule
             ToMP-50        & 34.7     & 17.4      & 7.0       & 3.4   \\ 
             TaMOs-50       & 19.2     & 17.9      & 16.3      & 13.9     \\
            \bottomrule
        \end{tabular}
	}%
    \vspace{-3mm}%
    \label{tab:run_time}%
\end{table}
\begin{table}[!t]
	\centering
	\vspace{0mm}
	\newcommand{\best}[1]{\textbf{\textcolor{red}{#1}}}
	\newcommand{\scnd}[1]{\textbf{\textcolor{blue}{#1}}}
	\newcommand{\opt}[1]{\textbf{\textcolor{violet}{#1}}}
	\newcommand{\fast}[1]{\textbf{\textcolor{orange}{#1}}}
	\newcommand{\dist}{\hspace{3pt}}%
	\caption{Comparison of GOT and MOT metrics on LaGOT.}%
        \vspace{-3mm}
	\resizebox{1.00\linewidth}{!}{%
        \begin{tabular}{l@{\dist}l@{\dist}c@{\dist}c@{\dist}c@{\dist}c@{\dist}c@{\dist}c@{\dist}c@{\dist}|c@{\dist}c@{\dist}c@{\dist}c@{\dist}}
        	\toprule
                &                    & F1-Score & Success & HOTA & MOTA & IDF1 & OWTA \\
                \midrule
                \multirow{2}{*}{GOT} & \textbf{TaMOs-SwinBase} & \best{0.643} & \best{63.1} & \best{62.1} & \best{58.2} & \best{74.7} & \best{68.9} \\
                                     & \textbf{TaMOs-50}       & \scnd{0.628} & 61.3        & 60.0        & \scnd{52.9} & 72.0        & 67.1        \\
                \midrule
                \multirow{6}{*}{SOT} & MixFormerLarge-22k      & 0.619        & \scnd{62.4} & \scnd{61.5} & 52.3        & \scnd{74.3} & \scnd{69.0} \\
                                     & ToMP-101                & 0.617        & 61.6        & 60.1        & 51.9        & 73.8        & 67.5        \\
                                     & STARK-101               & 0.603        & 60.3        & 59.4        & 49.0        & 72.5        & 67.0        \\
                                     & TransT                  & 0.605        & 59.5        & 57.7        & 46.6        & 70.7        & 65.6        \\
                                     & KeepTrack               & 0.626        & 61.3        & 59.1        & 51.3        & 73.8        & 66.2        \\
                                     & SuperDiMP               & 0.575        & 58.1        & 56.1        & 43.2        & 69.7        & 63.8        \\
                \midrule
                \multirow{2}{*}{MOT} & QDTrack                 & 0.187        & 19.2        & 22.2        & -115.8      & 16.3        & 36.3        \\
                                     & OVTrack                 & 0.128        & 13.4        & 24.4        & 13.9        & 23.5        & 25.9         \\
                \bottomrule
        \end{tabular}
	}\vspace{-6mm}
	\label{tab:metrics_lagot}%
\end{table}

\parsection{Results}
QDTrack and OVTrack achieve a VOTLT F1-Score of 0.187 and 0.128 respectively, performing inferior to all other trackers. Neither of the MOT trackers is robust enough and both fail to track rare or unknown generic objects.
To further explore the limitations of \gls{mot} methods in our setting, we evaluate ‘Oracle’ versions, where we select the track ID that maximizes the scores on LaGOT. 
Even with such oracle information, the performance of QDTrack and OVtrack is by far inferior to any evaluated \gls{sot} baseline (VOTLT 33.1 and 23.0 respectively).
In addition we evaluate both trackers using all its predicted tracks with \gls{mot} metrics, see Tab.~\ref{tab:metrics_lagot}. QDTrack tracks multiple background objects that are not annotated in LaGOT leading to many \glspl{fp}, and OVTrack tracks unannotated objects as well since the videos are not annotated exhaustively on class levels. Thus, traditional \gls{mot} tracking metrics such as MOTA, HOTA and IDF1 are unsuitable to evaluate MOT trackers on LaGOT. 
Instead, we concentrate on the OWTA metric~\cite{Liu_2022_CVPR_OpenWorldTracking} that focuses on \acrlong{detre} and \acrlong{assa} and thus ignores \glspl{fp}.
QDTrack achieves 36.3 and OVTrack 25.9, which are still the lowest OWTA scores compared to SOT and GOT trackers. 

\subsection{State-of-the-Art Comparison on SOT Datasets}\label{subsec:sot_experiments}
While TaMOs is built to track multiple objects in a video it can as well track only a single generic object. Thus, we evaluate TaMOs on popular large-scale \gls{sot} benchmarks. We deploy the very same tracker in these settings, without altering its weights or any hyper-parameters. 

\parsection{LaSOT~\cite{Fan_2019_CVPR_Lasot}}
This large-scale dataset consists of 280 test sequences with 2500 frames on average.  Tab.~\ref{tab:sot} shows a comparison to recent \gls{sot} trackers. While primarily designed to cope with multiple objects, our tracker achieves the highest precision and the third highest success rate \gls{auc}. Note, that neither MixFormer, SwinTrack nor OSTrack operate on the entire video frame, but rely on a local search area to produce such high tracking accuracy. 

\parsection{TrackingNet~\cite{2018_Muller_Trackingnet}}
This dataset consists of 511 test sequences and predictions are evaluated on a server.
Tab.~\ref{tab:sot} shows that our tracker with SwinBase sets the new state of the art on TrackingNet in terms of success rate and precision \gls{auc}. Similarly, our tracker with ResNet-50 achieves the best results among all trackers using that backbone.

The results on both benchmarks show the great potential of applying trackers {\em globally} without motion priors, such as search area selection~\cite{Bhat_2019_ICCV_DIMP,Cui_2022_CVPR_Mixformer,Ye_2022_ECCV_OSTrack} or spatial windowing~\cite{Li_2018_CVPR_SiamRPN,Li_2019_CVPR_SiamRPN++}.

\subsection{Ablation Study}
\label{subsec:ablation}

The ablation experiments shown in Tabs.~\ref{tab:ablation_enc} and~\ref{tab:ablation_arch_inf} are performed before the final annotation verification step such that the results compared to the numbers above slightly differ.

\parsection{Generic Multiple Object Encoding}
Tab.~\ref{tab:ablation_enc} shows the effect of the Gaussian score map encoding, the LTRB bounding box encoding and the total number of object embeddings $m$ stored in the pool $E$.
The first two rows in Tab.~\ref{tab:ablation_enc} show that the LTRB encoding is more important than the Gaussian encoding (as removing LTRB decreases all results more significantly).
Another key factor is the number of different object embeddings, that sets an upper limit on the number of objects that can be tracked.
LaGOT requires at least 10 embeddings and our tracker achieves the best results when using a pool size of 10. Increasing the number of embeddings decreases the overall tracking performance. 

\parsection{Architecture}
Tab.~\ref{tab:ablation_arch_inf} shows that using SwinBase increases the tracking performance on LaSOT and LaGOT.
Similarly, adding an \gls{fpn} improves  the results.

\parsection{Inference} During inference we update the memory by adding a second dynamic training frame similar to ToMP~\cite{Mayer_2022_CVPR_ToMP}. Since the ground truth bounding boxes are not available, we use the predicted boxes as annotations. We replace the dynamic training frame (update the memory) if the maximal
value in each target score map is above the threshold of $\tau = 0.85$. The results in Tab.~\ref{tab:ablation_arch_inf} show that adding a second training frame improves the results on both datasets.

\begin{table}[!t]
    \centering
    \newcommand{\best}[1]{\textbf{\textcolor{red}{#1}}}
    \newcommand{\scnd}[1]{\textbf{\textcolor{blue}{#1}}}
    \newcommand{\dist}{~~~}%
    \newcommand{\yes}{\textcolor{black}{\checkmark}}
    \newcommand{\no}{\textcolor{black}{\ding{55}}}
    \caption{State-of-the-art comparison on SOT datasets.}
    \vspace{-3mm}
    \resizebox{\linewidth}{!}{%
        \begin{tabular}{l@{\dist}c@{~~}|@{~~}c@{\dist}c@{\dist}c@{~~}|@{~~}c@{\dist}c@{\dist}c@{\dist}}
            \toprule
                                                            &         & \multicolumn{3}{c}{LaSOT~\cite{Fan_2019_CVPR_Lasot}} & \multicolumn{3}{c}{TrackingNet~\cite{2018_Muller_Trackingnet}}    \\
            Method                                          & Venue   & Prec & N-Prec & Succ & Prec & N-Prec & Succ    \\
            \midrule
            \textbf{TaMOs-SwinBase}                         & WACV'24 & \best{77.8} & 79.3        & 70.2  & \best{84.2}&\scnd{88.7} & \best{84.4}   \\
            \textbf{TaMOs-50}                               & WACV'24 & 75.0        & 77.2        & 67.9         & 82.0       & 87.2       & 82.7  \\
            \midrule
            SwinTrack~\cite{Lin_2021_NeurIPS_SwinTrack}     & NIPS'22 & 76.5        & ---         & \best{71.3}  & 82.0       & ---        & \scnd{84.0}  \\
            Unicorn~\cite{Bin_2022_ECCV_Unicorn}            & ECCV'22 & 74.1        & 76.6        & 68.5         & 82.2       & 86.4       & 83.0  \\
            AiATrack~\cite{Gao_2022_ECCV_AiATrack}          & ECCV'22 & 73.8        & 79.4        & 69.0         & 80.4       & 87.8       & 82.7  \\
            OSTrack~\cite{Ye_2022_ECCV_OSTrack}             & ECCV'22 & \scnd{77.6} & \best{81.1} & \scnd{71.1}  & \scnd{83.2}& 88.5       & \scnd{83.9}  \\
            RTS~\cite{Paul_2022_ECCV_RTS}                   & ECCV'22 & 73.7        & 76.2        & 69.7         & 79.4       & 86.0       & 81.6  \\
            MixFormer~\cite{Cui_2022_CVPR_Mixformer}        & CVPR'22 & 76.3        & \scnd{79.9} & 70.1         & 83.1       & \best{88.9}& 83.9  \\
            ToMP~\cite{Mayer_2022_CVPR_ToMP}                & CVPR'22 & 73.5        & 79.2        & 68.5         & 78.9       & 86.4       & 81.5  \\
            UTT~\cite{Ma_2022_CVPR_UTT}                     & CVPR'22 & 67.2        & ---         & 64.6         & 77.0       & ---        & 79.7  \\
            KeepTrack~\cite{Mayer_2021_ICCV_KeepTrack}      & ICCV'21 & 70.2        & 77.2        & 67.1         & 73.8       & 83.5       & 78.1  \\
            STARK~\cite{Yan_2021_ICCV_STARK}                & ICCV'21 & 72.2        & 77.0        & 67.1         & ---        & 86.9       & 82.0  \\
            TransT~\cite{Chen_2021_CVPR_TransT}             & CVPR'21 & 69.0        & 73.8        & 64.9         & 80.3       & 86.7       & 81.4  \\
            SuperDiMP~\cite{Danelljan_2020_CVPR_PRDIMP}     & CVPR'20 & 65.3        & 72.2        & 63.1         & 73.3       & 83.5       & 78.1  \\

            \bottomrule
        \end{tabular}
	}%
    \vspace{-3mm}
    \label{tab:sot}%
\end{table}
\begin{table}[!t]
    \centering
    \newcommand{\best}[1]{\textbf{\textcolor{red}{#1}}}
    \newcommand{\scnd}[1]{\textbf{\textcolor{blue}{#1}}}
    \newcommand{\dist}{\hspace{10pt}}%
    \newcommand{\yes}{\textcolor{black}{\checkmark}}
    \newcommand{\no}{\textcolor{black}{\ding{55}}}
    \caption{Analysis of different object encoding settings. All tested configurations are not employing the \acrshort{fpn}.}
    \vspace{-3mm}
    \resizebox{1\columnwidth}{!}{%
        \begin{tabular}{c@{\dist}c@{\dist}c@{\dist}c@{\dist}c@{\dist}c@{\dist}}
            \toprule
            Gaussian  & LTRB     & Object Embedding    & LaSOT       & \multicolumn{2}{c@{\dist}}{LaGOT}              \\
            Encoding  & Encoding & Pool size $m$       & AUC         & AUC         & F1 \\
            \midrule
            \yes      & \no      & 10                  & 58.3        & 54.0        & 0.552\\
            \no       & \yes     & 10                  & 66.3        & 60.2        & 0.620\\
            \yes      & \yes     & 10                  & \best{67.2} & \best{61.6} & \best{0.633}\\
            \yes      & \yes     & 15                  & 65.7        & 60.0        & 0.617\\
            \yes      & \yes     & 20                  & 65.7        & 58.9        & 0.603\\
            \yes      & \yes     & 50                  & 63.1        & 57.4        & 0.587\\
            \bottomrule
        \end{tabular}
	}%
    \vspace{-4mm}
    \label{tab:ablation_enc}%
\end{table}
\begin{table}[!t]
    \centering
    \newcommand{\best}[1]{\textbf{\textcolor{red}{#1}}}
    \newcommand{\scnd}[1]{\textbf{\textcolor{blue}{#1}}}
    \newcommand{\dist}{\hspace{20pt}}%
    \newcommand{\yes}{\textcolor{black}{\checkmark}}
    \newcommand{\no}{\textcolor{black}{\ding{55}}}
    \caption{Architecture and memory update analysis.}
    \vspace{-3mm}
    \resizebox{0.97\columnwidth}{!}{%
        \begin{tabular}{c@{\dist}c@{\dist}c@{\dist}c@{\dist}c@{\dist}c@{\dist}}
            \toprule
                      &          & Memory & LaSOT       & \multicolumn{2}{c@{\dist}}{LaGOT} \\
            Backbone  & FPN      & Update & AUC         & AUC         & F1 \\
            \midrule
            Resnet-50 & \no      & \yes   & 67.2        & 60.4        & 0.621 \\
            Resnet-50 & \yes     & \no    & 66.0        & 60.2        & 0.620 \\
            Resnet-50 & \yes     & \yes   & \best{67.9} & \best{61.6} & \best{0.633} \\
            \midrule
            SwinBase  & \no      & \yes   & 69.5        & 62.4        & 0.643 \\
            SwinBase  & \yes     & \no    & 67.9        & 62.1        & 0.636 \\
            SwinBase  & \yes     & \yes   & \best{70.2} & \best{63.5} & \best{0.649} \\
            \bottomrule
        \end{tabular}
	}%
    \vspace{-7mm}
    \label{tab:ablation_arch_inf}%
\end{table}
\section{Conclusion}

We propose a novel multiple object \gls{got} tracking benchmark, LaGOT, that allows to evaluate \gls{got} methods that can jointly track multiple targets in the same sequence. We demonstrate that the proposed task and benchmark are challenging for existing SOT and MOT trackers.
We further propose a Transformer-based tracker capable of processing multiple targets at the same time, with a novel generic multi object encoding and an \gls{fpn} 
in order to achieve full frame tracking. Our method outperforms recent trackers on the LaGOT benchmark, while operating $4\times$ faster than the \gls{sot} baseline when tracking 10 objects. Lastly, our approach also achieves excellent results on popular \gls{sot} benchmarks.

{\noindent\textbf{Funding:}
This work was done at Google Research.}

{\noindent\textbf{Acknowledgement:}
The authors thank Paul Voigtlaender for his support, fruitful discussions and valuable feedback.}

%%%%%%%%% REFERENCES
{\small
\bibliographystyle{ieee_fullname}
\bibliography{egbib}
}

\begin{appendices}
    In this supplementary material, we first give an overview of the different task definitions and the corresponding abbreviations used in the paper and supplementary material. Next, we describe the details of the model architecture and training in Sec.~\ref{sup:sec:training-inference}. Then, we provide more insights into the experiments presented in the main paper and provide additional results on less popular tracking datasets in Sec.~\ref{sup:sec:experiments}. Then, we show visual results between the baseline and our tracker on multiple sequences of the proposed datasets including failure cases~\ref{sup:sec:visual-results}. Next, we discuss the limitations of the proposed tracker in Sec.~\ref{sup:sec:limitations}. Finally, we provide additional insights about our dataset and compare it to datasets of related tasks in Sec.~\ref{sup:sec:datasets}.

\section{Glossary}
In this Section we will briefly summarize the different task definitions behind the individual abbreviations:

\parsection{GOT} Generic Object Tracking refers to the task of tracking potentially multiple user-defined target objects of arbitrary classes specified by a user-specified bounding box in the initial video frame.\\
\parsection{SOT} Singe Object Tracking is the same task as GOT but focuses on the setting where only a single generic object needs to be tracked.\\
\parsection{Multi-Object GOT} The same as GOT but emphasizes that multiple-objects need to be tracked. We use multi-object GOT because GOT is in other research works sometimes used interchangeably with SOT.\\
\parsection{MOT} Multi Object Tracking is completely different from the tasks listed above because it requires a class category list to detect and track all objects corresponding to the defined class categories.\\
\parsection{GMOT} Generic Multi Object Tracking is the same as MOT but instead of using a class category list to define the target objects, a single user-specified box shows an example object of the target class category. Thus, all objects that belong to the same class as the user-specified example need to be detected and tracked.

\section{Model Architecture and Training Details}\label{sup:sec:training-inference}
\parsection{Architecture} We extract backbone features either from the ResNet-50 or the SwinBase backbone. For both backbones we extract the features corresponding to the blocks with stride 8 and 16. We only use the features with stride 16 for object encoding and  feed these features into the model predictor. For both backbones we use a linear layer to decrease the number of channels from 1024 to 256 or 512 to 256 respectively. Thus we use 256 dimensional object embeddings $e_i$ and a \gls{mlp} to project the LTRB bounding box encoding map from 4 to 256 channels. Since the model predictor produces 256 dimensional convolutional filters we require the same number of channels for the \gls{fpn} output features. In particular we use a two layer \gls{fpn} that uses as input the enhanced Transformer encoder output features corresponding to the test frame as well as the aforementioned high resolution backbone test features. The high resolution input features have either 512 or 256 channels for the Resnet-50 or the SwinBase backbone respectively. Thus, we adapt the \gls{fpn} accordingly depending on the used backbone.
\begin{figure*}[t]
\centering
  \subfloat[Success Plot]{
    \includegraphics[width=\columnwidth, keepaspectratio]{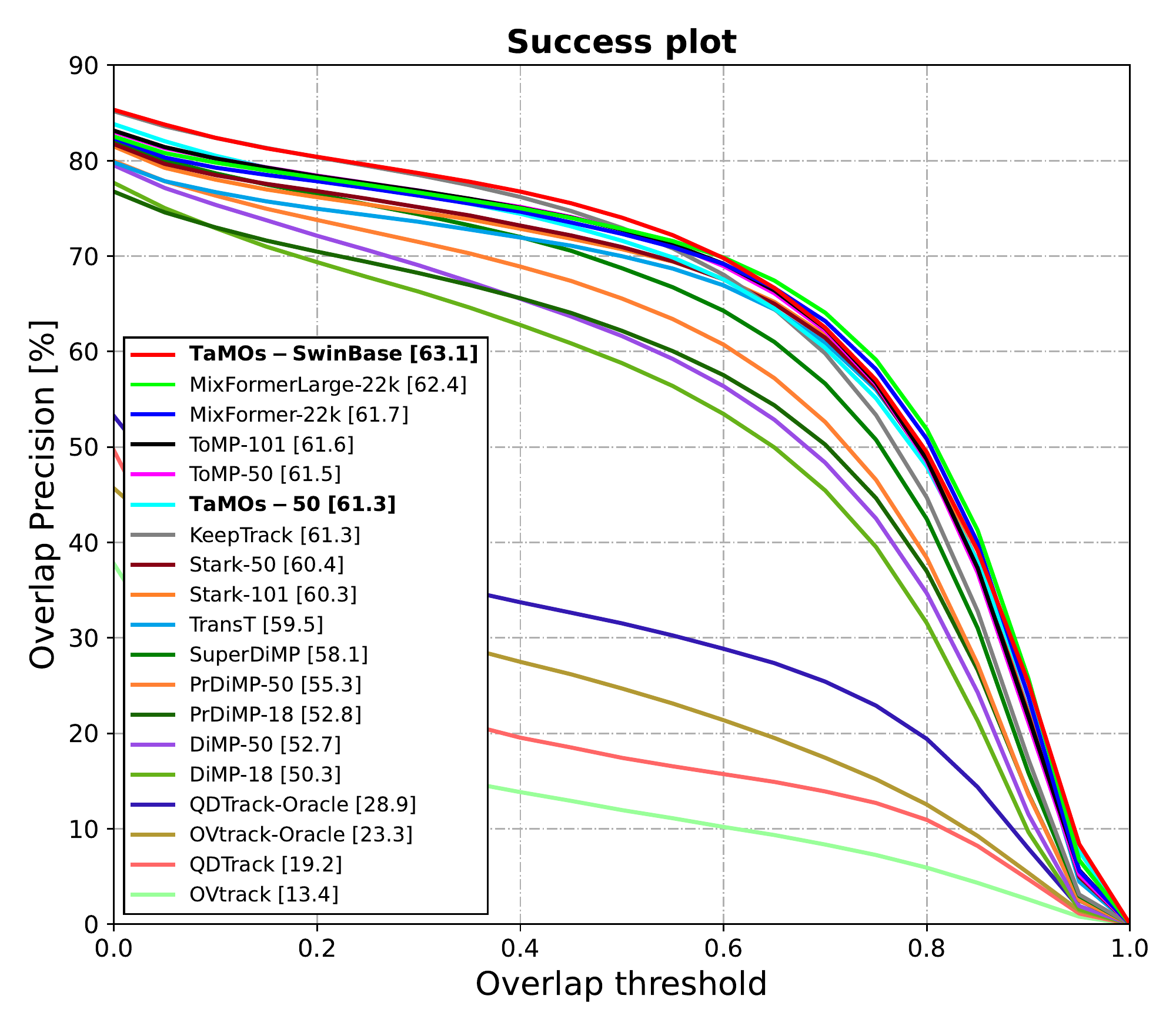}
    \label{sup:fig:success}
  }
  \hfill
  \subfloat[Precision-Recall]{
    \includegraphics[width=\columnwidth, keepaspectratio]{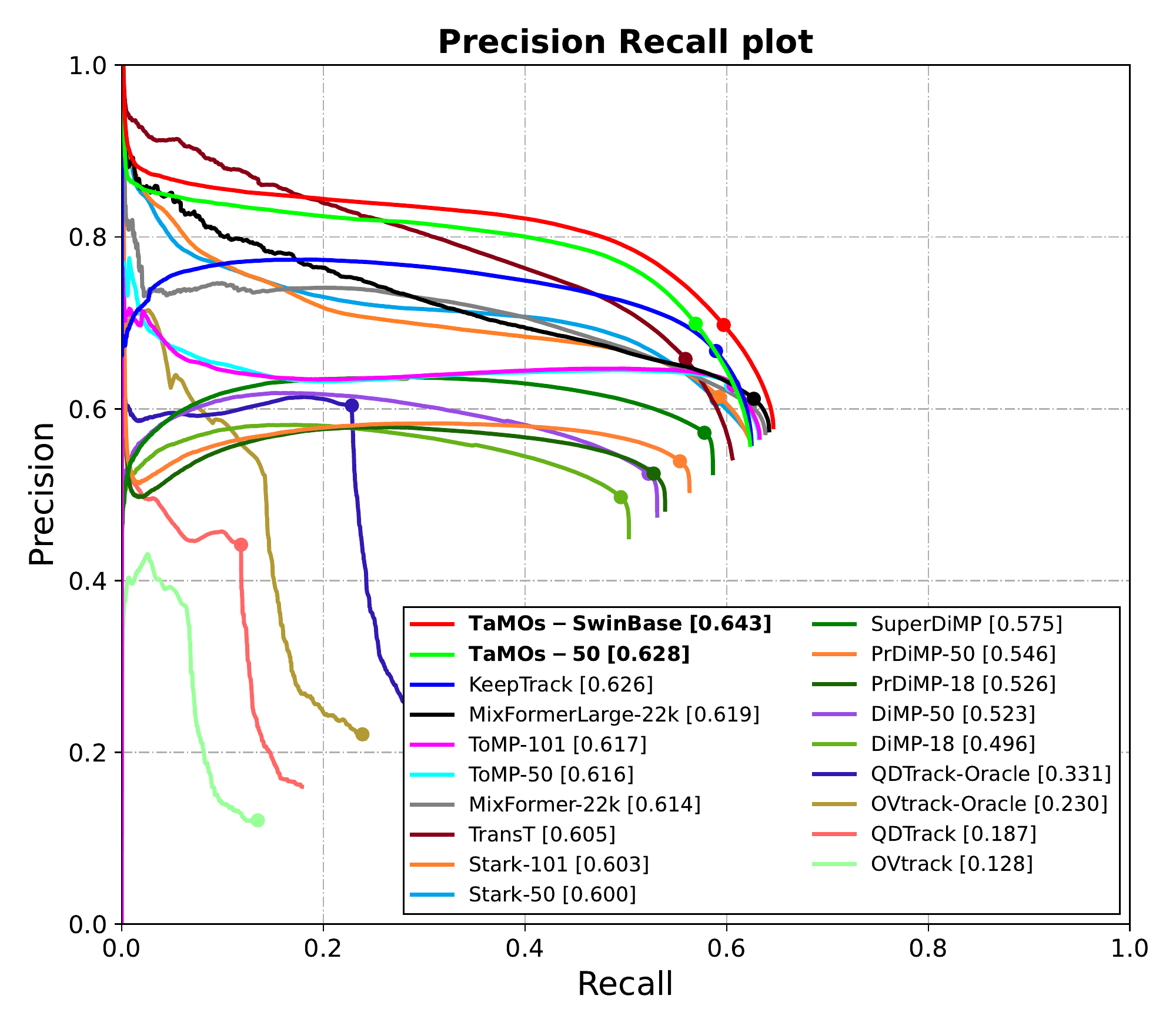}
    \label{sup:fig:prec_rec}
  }
\caption{Success plot, showing $\text{OP}_T$, on LaGOT (AUC is reported in the legend). Tracking Precision-Recall curve on LaGOT -- VOTLT  is reported in the legend (the highest F1-score).}\label{sup:fig:lagot_results}
\end{figure*}

\parsection{Training Details}
Since our tracker operates on full frames, we retain the full training and testing frames. 
The frames are re-scaled and padded to a resolution of $384\times576$. 
As we use the feature maps with stride 16 for both the ResNet-50~\cite{He_2016_CVPR_Resnet} and SwinBase~\cite{Liu_2021_ICCV_Swin} backbones, this results in an extracted feature and score map resolution of $24\times36$. 
For ResNet-50 we use pretrained weights on ImageNet-1k and for SwinBase on ImageNet-22k.
We use a fixed size Gaussian when producing the score map encoding for each object where $\sigma = 0.25$. 
Furthermore, we use gradient norm clipping with the parameter 0.1 in order to stabilize training. In addition, we employ data augmentation techniques during training such as random scaling and cropping in addition to color jittering and randomly flipping the frame.
The regression loss is given by
\begin{equation}
     L_\mathrm{bbreg} = \sum_{i=0}^{n}L_\mathrm{GIoU}\left(\hat{b}_i^\mathrm{ltrb}, \hat{b}_i^\mathrm{ltrb}\right),
\end{equation}
where $L_\mathrm{GIoU}$ denotes the generalized IoU-Loss~\cite{Rezatofighi_2019_CVPR_GIOU}.
The overall training loss is then defined as
\begin{equation}
    L_\mathrm{tot} = \lambda_\mathrm{cls}L_\mathrm{cls}(\hat{y}, y) + \lambda_\mathrm{bbreg}\cdot L_\mathrm{bbreg}(\hat{b}^\mathrm{ltrb}, b^\mathrm{ltrb})
\end{equation}
where $\lambda_\mathrm{cls}= 100$ and $\lambda_\mathrm{bbreg}= 1$ are scalars weighting the contribution of each loss component.
We use ADAMW~\cite{Loshchilov_2019_ICLR_ADAMW} with a learning rate of 0.0001 that we decay after 150 and 250 epochs by a factor of 0.2 and
train all models on four A100 GPUs with a batch size of $4 \times 12$ or $4 \times 6$.

\parsection{Inference}
During inference we adopt the simple memory updating approach described in~\cite{Mayer_2022_CVPR_ToMP}. In particular, updating the memory refers to adding a second dynamic training frame using predicted box annotations. We replace the second training frame (update the memory) if the maximal value in each score map is above the threshold of $\tau = 0.85$.

For accurate bounding box prediction and localization we employed an \gls{fpn}.
In contrast to training, where we applied the target models directly on the Transformer encoder features and also on the low- and high-resolution \gls{fpn} feature maps, we only use the high-resolution score and bounding box prediction maps during inference. We empirically observed better training performance when applying the losses on each instead of only on the high resolution outputs. However, during inference we are only interested in the high resolution predictions.

\begin{table}[!t]
	\centering
	\vspace{0mm}
	\newcommand{\best}[1]{\textbf{\textcolor{red}{#1}}}
	\newcommand{\scnd}[1]{\textbf{\textcolor{blue}{#1}}}
	\newcommand{\opt}[1]{\textbf{\textcolor{violet}{#1}}}
	\newcommand{\fast}[1]{\textbf{\textcolor{orange}{#1}}}
	\newcommand{\dist}{\hspace{3pt}}%
	\caption{Comparison of the combination of GOT and MOT methods. GOT return the detections and the MOT methods are used for object association over time on LaGOT.}%
	\resizebox{1.00\linewidth}{!}{%
        \begin{tabular}{l@{\dist}|l@{\dist}|c@{\dist}c@{\dist}|c@{\dist}c@{\dist}c@{\dist}c@{\dist}c@{\dist}|c@{\dist}c@{\dist}c@{\dist}c@{\dist}}
        	\toprule
                GOT                                 & MOT         & F1-Score & Success & HOTA & MOTA & IDF1 \\
                \midrule
                \multirow{3}{*}{TaMOs-SwinBase}     & ---         & 0.643    & 63.1    & 62.1 & 58.2 & 74.7 \\
                                                    & SORT~\cite{Bewley_2016_ICIP_SORT}        & 0.438    & 35.7    & 45.9 & 52.2 & 43.3 \\
                                                    & ByteTrack~\cite{Zhang_2022_ECCV_ByteTrack}   & 0.459    & 37.7    & 50.4 & 57.1 & 53.9 \\
                \midrule
                \multirow{3}{*}{MixFormerLarge-22k} &  ---        & 0.619    & 62.4    & 61.5 & 52.3 & 74.3 \\
                                                    &  SORT~\cite{Bewley_2016_ICIP_SORT}       & 0.418    & 34.0    & 45.6 & 43.9 & 44.9 \\
                                                    &  ByteTrack~\cite{Zhang_2022_ECCV_ByteTrack}  & 0.450    & 36.4    & 47.5 & 44.8 & 49.6 \\
                \bottomrule

        \end{tabular}
	}
	\label{sup:tab:lagot_got_mot}%
\end{table}
\begin{table*}[!t]
	\centering
	\vspace{0mm}
	\newcommand{\best}[1]{\textbf{\textcolor{red}{#1}}}
	\newcommand{\scnd}[1]{\textbf{\textcolor{blue}{#1}}}
	\newcommand{\opt}[1]{\textbf{\textcolor{violet}{#1}}}
	\newcommand{\fast}[1]{\textbf{\textcolor{orange}{#1}}}
	\newcommand{\dist}{\hspace{15pt}}%
	\caption{Comparison of different trackers using MOT metrics on LaGOT.}%
	\resizebox{1.00\linewidth}{!}{%
        \begin{tabular}{l@{\dist}l@{\dist}c@{\dist}c@{\dist}c@{\dist}c@{\dist}c@{\dist}c@{\dist}c@{\dist}c@{\dist}c@{\dist}c@{\dist}c@{\dist}c@{\dist}c@{\dist}}
            \toprule
            & & HOTA & DetA & AssA & DetRe & DetPr & AssRe & AssPr & LocA & OWTA & MOTA & IDSW & IDF1 \\
            \midrule
            \multirow{2}{*}{GOT}    & \textbf{TaMOs-SwinBase} & 62.1 & 57.3 & 68.4 & 69.9 & 69.9 & 75.9 & 75.9 & 84.2 & 68.9 & 58.2 & 6734 & 74.7 \\
                                    & \textbf{TaMOs-50}       & 60.0 & 54.6 & 66.9 & 67.7 & 67.7 & 74.5 & 74.5 & 84.0 & 67.1 & 52.9 & 7901 & 72.0 \\
            \midrule
            \multirow{13}{*}{SOT}   & MixformerLarge-22k      & 61.5 & 53.8 & 70.9 & 67.4 & 67.4 & 77.8 & 77.8 & 84.8 & 69.0 & 52.3  & 3150 & 74.3 \\
                                    & Mixformer-22k           & 61.2 & 54.0 & 70.0 & 67.4 & 67.4 & 77.0 & 77.0 & 84.5 & 68.6 & 53.2  & 3339 & 74.4 \\
                                    & ToMP-101                & 60.1 & 53.0 & 68.8 & 66.4 & 66.4 & 76.2 & 76.2 & 83.9 & 67.5 & 51.9  & 2638 & 73.8 \\
                                    & ToMP-50                 & 60.0 & 53.0 & 68.6 & 66.4 & 66.4 & 76.0 & 76.1 & 83.8 & 67.4 & 52.3  & 2378 & 74.0 \\
                                    & STARK-ST-101            & 59.4 & 51.8 & 68.8 & 65.6 & 65.6 & 75.9 & 75.9 & 84.2 & 67.1 & 49.0  & 3568 & 72.5 \\
                                    & STARK-ST-50             & 59.4 & 51.9 & 68.5 & 65.6 & 65.6 & 75.6 & 75.6 & 83.9 & 66.9 & 49.5  & 4277 & 72.6 \\
                                    & TransT                  & 57.8 & 50.2 & 67.1 & 64.3 & 64.3 & 74.5 & 74.6 & 84.3 & 65.6 & 46.6  & 2323 & 70.7 \\
                                    & KeepTrack               & 59.1 & 52.3 & 67.3 & 65.4 & 65.4 & 74.7 & 74.7 & 82.3 & 66.2 & 51.3  & 2299 & 73.8  \\
                                    & SuperDiMP               & 56.1 & 48.3 & 65.8 & 62.1 & 62.1 & 73.5 & 73.5 & 82.2 & 63.8 & 43.2  & 1966 & 69.7 \\
                                    & PrDiMP-50               & 53.0 & 45.6 & 62.1 & 59.6 & 59.6 & 70.3 & 70.4 & 81.3 & 60.7 & 38.4  & 2380 & 66.6 \\
                                    & PrDiMP-18               & 51.4 & 42.8 & 62.2 & 57.2 & 57.2 & 70.2 & 70.3 & 81.3 & 59.5 & 31.9  & 1981 & 63.4 \\
                                    & DiMP-50                 & 50.8 & 42.1 & 62.0 & 56.2 & 56.2 & 69.7 & 69.7 & 80.2 & 58.9 & 29.4  & 1680 & 62.1 \\
                                    & DiMP-18                 & 48.1 & 39.3 & 59.6 & 53.5 & 53.5 & 67.6 & 67.6 & 79.5 & 56.3 & 23.2  & 1757 & 59.0 \\
            \midrule
            \multirow{2}{*}{MOT}    & QDTrack                 & 22.2 & 17.3 & 29.0 & 46.2 & 21.0 & 30.3 & 80.0 & 81.8 & 36.3 & -115.8 & 18521 & 16.3 \\
                                    & OVTrack                 & 24.4 & 20.3 & 29.9 & 22.7 & 59.7 & 31.2 & 78.2 & 82.0 & 25.9 & 13.9   & 4951  & 23.5 \\
            \bottomrule
            \end{tabular}
	}
	\label{sup:tab:mot_lagot}%
\end{table*}
\begin{figure*}[t]
\centering
\subfloat[LaSOT]{
    \includegraphics[width=\columnwidth, keepaspectratio]{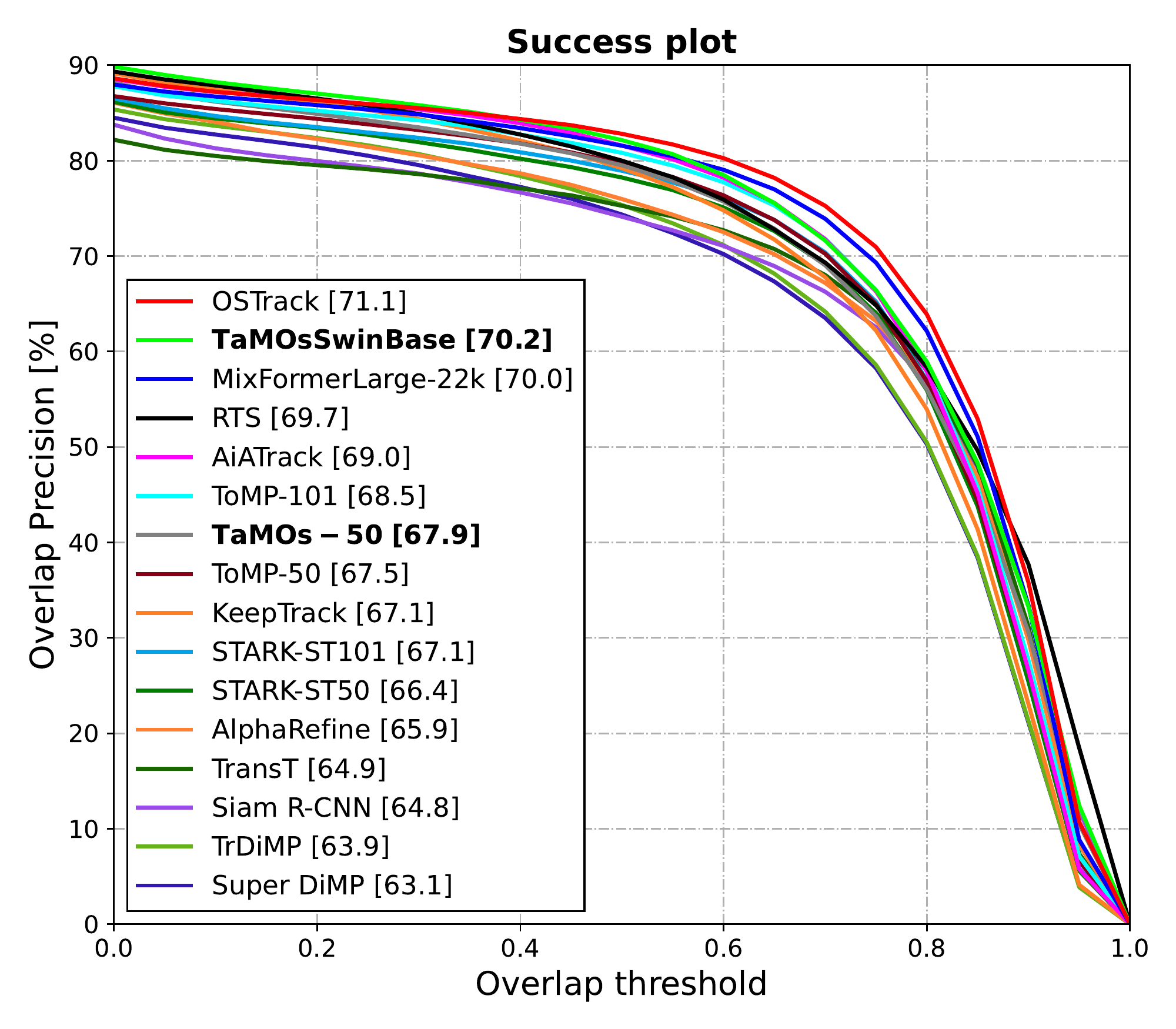}
    \label{sup:fig:lasot}
  }
  \hfill
  \subfloat[AViST]{
    \includegraphics[width=\columnwidth, keepaspectratio]{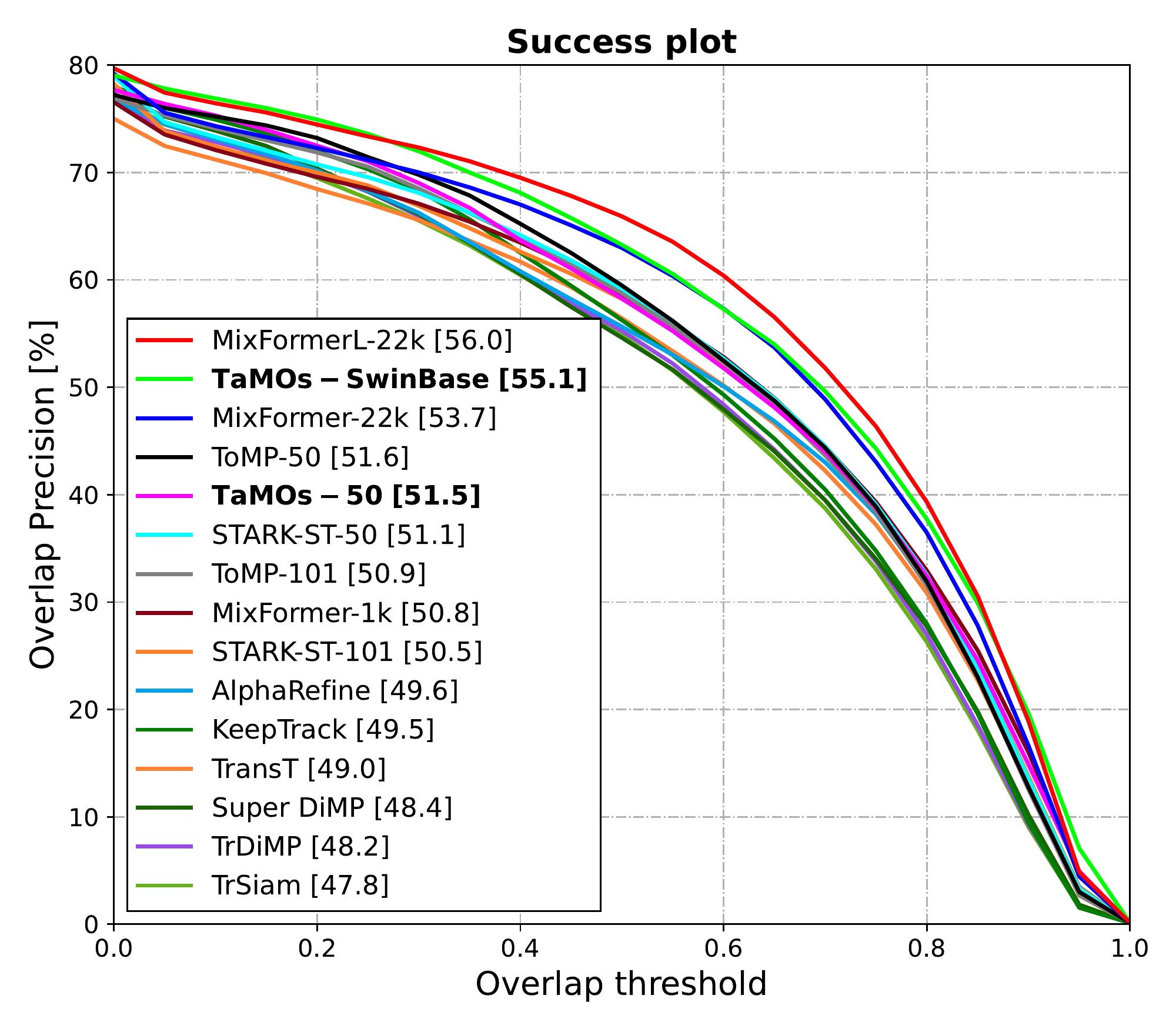}
    \label{sup:fig:avist}
  }
\caption{Success plot, showing $\text{OP}_T$, on LaSOT~\cite{Fan_2019_CVPR_Lasot} and AVisT~\cite{Noman_2022_BMVC_AVisT} (AUC is reported in the legend).}\label{sup:fig:success_plots}
\end{figure*}

\begin{table}[!t]
	\centering
	\vspace{-0mm}
	\newcommand{\best}[1]{\textbf{\textcolor{red}{#1}}}
	\newcommand{\scnd}[1]{\textbf{\textcolor{blue}{#1}}}
	\newcommand{\dist}{\hspace{15pt}}%
        \caption{Comparison to the state of the art on LaSOTExt~\cite{Fan_2020_IJCV_Lasot_ext}.}\label{sup:tab:lasotext}%
        \resizebox{\linewidth}{!}{%
        \begin{tabular}{l@{\dist}c@{\dist}|@{\dist}c@{\dist}c@{\dist}c@{\dist}}
            \toprule
                                                            &         & \multicolumn{3}{c}{LaSOTExt~\cite{Fan_2020_IJCV_Lasot_ext}}    \\
            Method                                          & Venue   & Prec        & N-Prec      & Succ     \\
            \midrule
            \textbf{TaMOs-SwinBase}                         &         & \best{58.0} & 57.8        & \scnd{49.2} \\
            \textbf{TaMOs-Resnet-50}                        &         & 54.1        & 55.0        & 46.7     \\
            \midrule
            AiATrack~\cite{Gao_2022_ECCV_AiATrack}          & ECCV'22 & 54.7        & 58.8        & 49.0     \\
            OSTrack~\cite{Ye_2022_ECCV_OSTrack}             & ECCV'22 & \scnd{57.6} & \scnd{61.3} & \best{50.5}     \\
            ToMP-101~\cite{Mayer_2022_CVPR_ToMP}            & CVPR'22 & 52.6        & 58.1        & 45.9     \\
            ToMP-50~\cite{Mayer_2022_CVPR_ToMP}             & CVPR'22 & 51.9        & 57.6        & 45.4     \\
            GTELT~\cite{Zhou_2022_CVPR_GTELT}               & CVPR'22 & 52.4        & 54.2        & 45.0     \\
            KeepTrack~\cite{Mayer_2021_ICCV_KeepTrack}      & ICCV'21 & 54.7        & \best{61.7} & 48.2     \\
            SuperDiMP~\cite{Danelljan_2020_CVPR_PRDIMP}     & CVPR'20 & 49.0        & 56.3        & 43.7     \\
            LTMU~\cite{Dai_2020_CVPR_LTMU}                  & CVPR'20 & 45.4        & 53.6        & 41.4     \\
            DiMP~\cite{Bhat_2019_ICCV_DIMP}                 & ICCV'19 & 43.2        & 49.6        & 39.2     \\
            ATOM~\cite{Danelljan_2019_CVPR_ATOM}            & CVPR'19 & 41.2        & 49.6        & 37.6     \\
            \bottomrule
        \end{tabular}
	}%
\end{table}

\section{Experiments}\label{sup:sec:experiments}
We provide more detailed results to complement the comparison shown in the main paper.
In addition we provide result for the LaSOTExt~\cite{Fan_2020_IJCV_Lasot_ext} dataset in order to assess the performance of our tracker on sequences containing small objects. Similarly, we analyze the capability of our tracker to handle adverse tracking conditions on AVisT~\cite{Noman_2022_BMVC_AVisT}. Furthermore, to provide results on another multiple object dataset we run the tracker on ImageNetVID~\cite{Russakovsky_2015_IJCV_ILSVRC15}. 
\subsection{LaGOT}
To complement the results shown in the main paper, we report in Fig.~\ref{sup:fig:lagot_results} and Tab.~\ref{sup:tab:mot_lagot} results for additional trackers and different variants, such as using a different backbone or different hyper-parameters. In Tab.~\ref{sup:tab:mot_lagot} we report additional \gls{mot} sub-metrics and statistics on LaGOT. In general we conclude, that using larger backbones especially if they are pretrained on ImageNet-22k leads to the best results. Furthermore, we observe that the \gls{mot} methods QDTrack and OVTrack (evaluated with default parameters provided in the OVTrack GitHub repository\footnote{https://github.com/SysCV/ovtrack}) are not competitive with \gls{got} methods. In particular, we observe that QDTrack and OVTrack achieve very low OWTA scores that depend on the \gls{detre} and the \gls{assa} scores. OVTrack scores the lowest \gls{detre} despite being an open-vocabulary detector. While this is an expected limitation, we further observe that QDTrack achieves by far the lowest \gls{assa} caused by the poor \gls{assre} of 30.3 compared to DiMP-18 that achieves 67.6.

In addition to the \gls{sot} and \gls{mot} baselines presented in the main paper, we also evaluate an open-world tracker~\cite{Liu_2022_CVPR_OpenWorldTracking}. Such a tracker aims at tracking all objects in the scene and should therefore also be able to track the generic objects contained in LaGOT.
In particular, we follow Liu~\etal~\cite{Liu_2022_CVPR_OpenWorldTracking} and generate object proposals for each video frame using their provided open-world detector. Then, we run SORT~\cite{Bewley_2016_ICIP_SORT} on top of the generated proposals using the default parameters. This leads to an OWTA score of 12.58, AssA of 3.57 and DetRe of 46.72.
We conclude that the complex videos with long tracks of the proposed benchmark are for now too challenging for existing open-world trackers.

Finally, we add another experiment where we use our tracker TaMOs or the \gls{sot} tracker Mixformer as one-shot object detectors and feed their detections and scores to a \gls{mot} tracker that focuses on building the final tracklets. In particular we use the popular SORT~\cite{Bewley_2016_ICIP_SORT} tracker and the recent state-of-the-art tracker ByteTrack~\cite{Zhang_2022_ECCV_ByteTrack}. For TaMOs and Mixformer, using their predicted bounding boxes and object ids leads to far better results than using an \gls{mot} method on top for post-processing. This behaviour holds when measuring the performance of the resulting trackers with \gls{got} as well as with \gls{mot} metrics, see Tab.~\ref{sup:tab:lagot_got_mot}.  While there is potential to increase the robustness of \gls{got} trackers in case of multiple objects, directly applying \gls{mot} trackers is not a good solution. Instead dedicated association algorithms for multi-object \gls{got} are needed. We conclude, that TaMOs and the proposed \gls{sot} trackers run in parallel, are solid baselines for LaGOT.

\subsection{LaSOT}
In addition to the result table, shown in the main paper, we show in Fig.~\ref{sup:fig:lasot} the success plot for LaSOT~\cite{Fan_2019_CVPR_Lasot}. We observe that our tracker is the most robust ($T< 0.3$). Furthermore, the plot shows that both MixFormerLarge-22k and OSTrack can regress more accurate bounding boxes ($0.5 < T < 0.9)$. However, unlike these specialized single-target object trackers, our approach is capable of jointly tracking multiple targets.
\begin{figure*}[p]
    \centering
    \includegraphics[width=\linewidth, keepaspectratio]{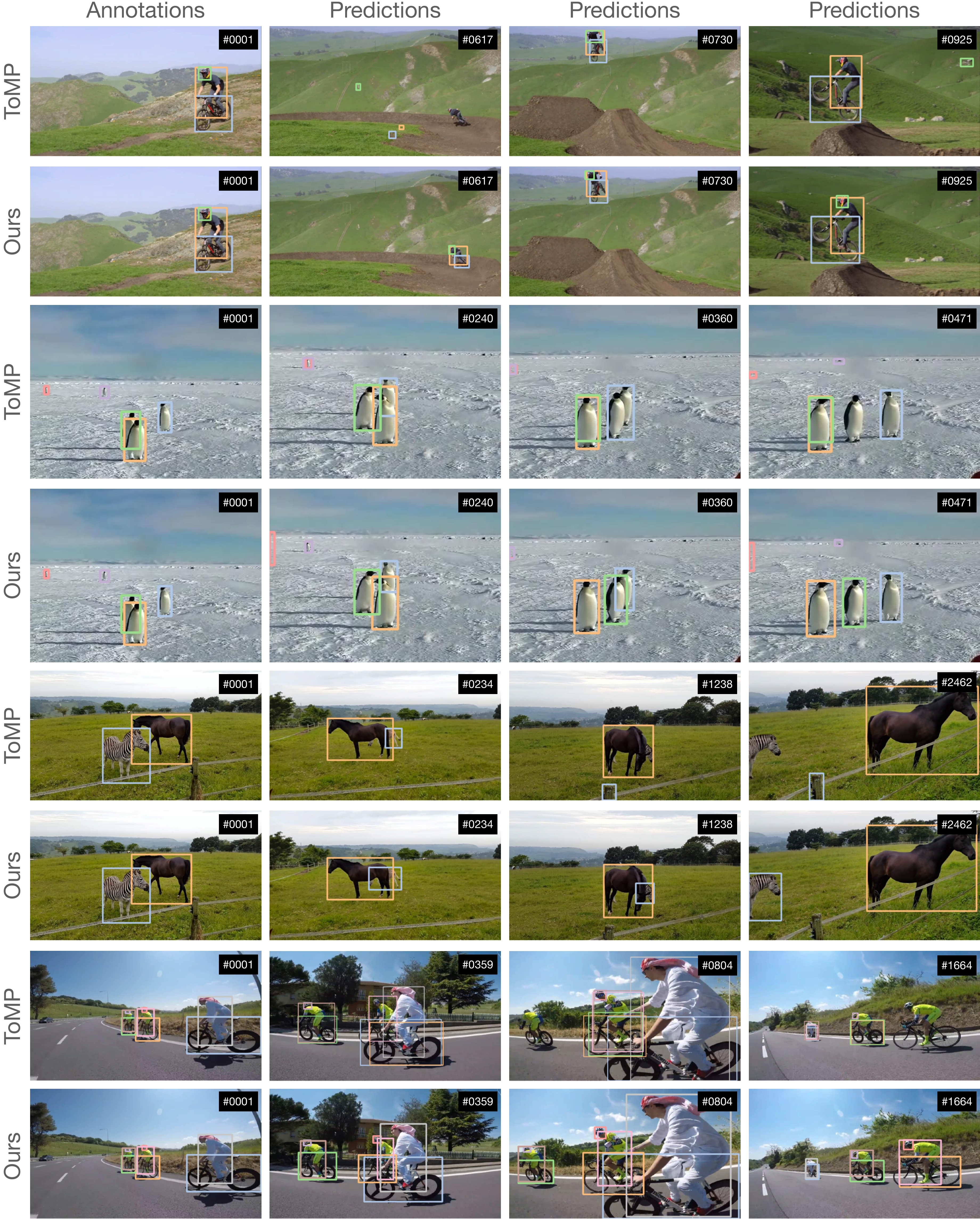}
\caption{Visual comparison between the proposed tracker (Ours-SwinBase) and the baseline ToMP-101 on different LaGOT sequences.}\label{sup:fig:visual_examples}
\end{figure*}
\begin{table}[!t]
    \centering
    \newcommand{\best}[1]{\textbf{\textcolor{red}{#1}}}
    \newcommand{\scnd}[1]{\textbf{\textcolor{blue}{#1}}}
    \newcommand{\dist}{\hspace{20pt}}%
    \newcommand{\yes}{\textcolor{black}{\checkmark}}
    \newcommand{\no}{\textcolor{black}{\ding{55}}}
    \caption{Analysis of the FPN and the zooming mechanism on LaSOTExt~\cite{Fan_2020_IJCV_Lasot_ext} and UAV123~\cite{Mueller_2016_ECCV_UAV123}.}%
    \resizebox{\columnwidth}{!}{%
        \begin{tabular}{c@{\dist}c@{\dist}c@{\dist}c@{\dist}c@{\dist}}
            \toprule
                      &         &        & LaSOTExt    & UAV123\\
            Backbone  & FPN     & Zoom   & AUC         & AUC\\
            \midrule
            Resnet-50 & \no     & \no    & 41.3        & 56.2 \\
            Resnet-50 & \yes    & \no    & 43.1        & 58.2 \\
            Resnet-50 & \yes    & \yes   & \best{46.7} & \best{64.2} \\
            \midrule
            SwinBase  & \no     & \no    & 43.9        & 56.5 \\
            SwinBase  & \yes    & \no    & 44.6        & 57.3\\
            SwinBase  & \yes    & \yes   & \best{49.2} & \best{66.2} \\
            \bottomrule
        \end{tabular}
	}%
    \label{sup:tab:ablation_zoom}%
\end{table}

\subsection{LaSOTExt}
Since our tracker always operates on the full frame without the help of a local search region, tracking small objects is challenging. Thus, we integrated an \gls{fpn} in our tracker to improve the tracking accuracy. To analyze our tracker on small objects we run it on LaSOTExt~\cite{Fan_2020_IJCV_Lasot_ext} and UAV123~\cite{Mueller_2016_ECCV_UAV123}. Tab.~\ref{sup:tab:ablation_zoom} shows that including an \gls{fpn} improves the tracking results on both datasets but is more effective when using a Resnet-50 as backbone. 

To track small objects a high feature map resolution is desirable. To better cope with extremely small objects, found in some \gls{sot} benchmarks, we add a simple zooming mechanism. In particular, when the target is smaller that $30\times30$ pixels, we crop a region of the image that ensures this minimal target size when up-scaled to the input-resolution of $384\times 576$.
Tab.~\ref{sup:tab:ablation_zoom} clearly shows that using such a zooming mechanism improves the results on LaSOTExt and UAV123 considerably, due to the presence of extremely small objects in these datasets. 

Tab.~\ref{sup:tab:lasotext} shows that our tracker with \gls{fpn} and zooming achieves competitive results on LaSOTExt. In particular it achieves the highest precision and the second highest success AUC only being outperformed by OSTrack~\cite{Ye_2022_ECCV_OSTrack}.

\subsection{AVisT}
In order to validate our tracker in adverse visibility scenarios we run it on AVisT~\cite{Noman_2022_BMVC_AVisT}. Fig~\ref{sup:fig:avist} shows that our tracker achieves excellent results with a success AUC of 55.1. This result shows that our tracker is able to track generic single objects even in visually challenging scenarios. The best tracker MixFormerLarge-22k is able to regress more accurate bounding boxes ($0.3 < T < 0.9$), as it relies on small search area selection to ensure high-resolution features. In contrast, our approach is capable of jointly tracking multiple objects. 

\subsection{ImageNetVID}
\begin{figure}[t]
    \centering
    \includegraphics[width=\columnwidth, keepaspectratio]{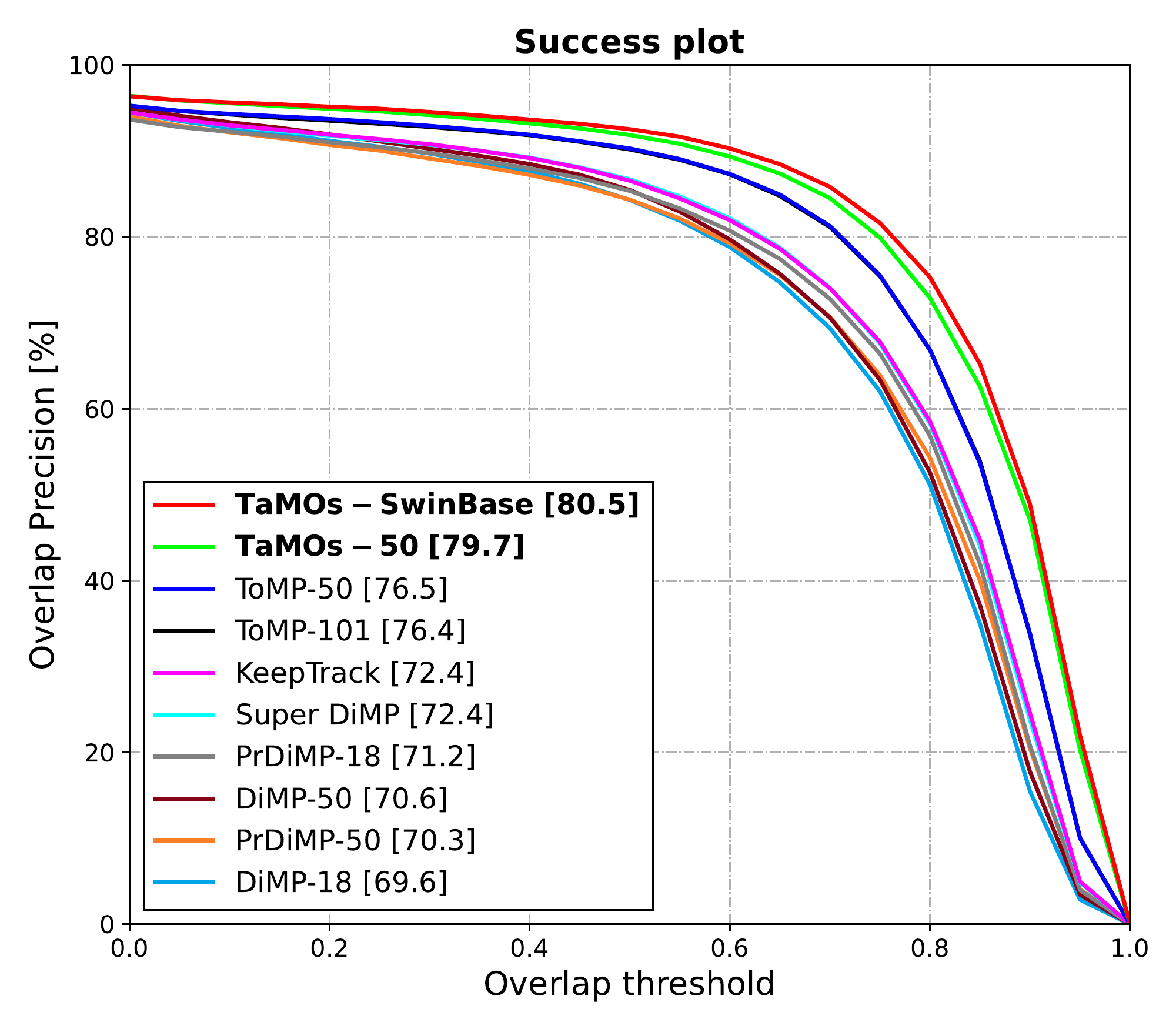}
\caption{Success plot, showing $\text{OP}_T$, on ImagenetVID~\cite{Russakovsky_2015_IJCV_ILSVRC15} (AUC is reported in the legend).}\label{sup:fig:imagenetvid_success_plots}
\end{figure}
In order to validate the proposed multiple object GOT tracker not only on LaGOT but also on another multiple object dataset, we modify  ImageNetVID~\cite{Russakovsky_2015_IJCV_ILSVRC15}. Since ImageNetVID is a video object detection datasets instead of a GOT dataset we perform the following adaptations. First, we remove all tracks that are not present in the first frame. Then, we use the remaining tracks to produce the bounding box annotations of the first frame. For simplicity we remove the 11 sequence where no track is visible in the first frame. This results in 544 sequences with 938 tracks and 1.7 tracks on average per video. Fig.~\ref{sup:fig:imagenetvid_success_plots} shows the success plot on the resulting multiple object GOT dataset. We observe that all trackers achieve relatively high AUC mostly differing in bounding box accuracy. Both versions of our tracker outperform the baselines ToMP-50 and ToMP-101~\cite{Mayer_2022_CVPR_ToMP}. In particular, we notice the superior bounding box accuracy of our tracker compared to ToMP. To summarize we observe a similar ranking between trackers on ImageNetVID and the proposed LaGOT dataset. However, LaGOT is more challenging due to the higher average track number (2.9 vs. 1.7) and the much longer sequence length (2258 vs.\ 312) that leads more frequently to occlusions and out-of-view events.

\section{Visual Results}\label{sup:sec:visual-results}
\parsection{Visual Comparison to the State of the Art}
We show visualizations of the tracking results of the baseline (ToMP-101) and our proposed tracker (TaMOs-SwinBase) on four different sequences of the proposed LaGOT benchmark in Fig.~\ref{sup:fig:visual_examples}. The first frame specifies the target objects annotated with bounding boxes that should be tracked in the video. The other frames show predictions of both trackers. The results on the first and third sequences demonstrate that our tracker can re-detect occluded objects quickly whereas a search area based tracker is not able to re-detect the targets if they reappear outside of the search area.
The second and fourth sequences show the superior robustness of our tracker. It is able to distinguish similarly looking objects better without confusing their ids. For more visual results we refer the reader to the mp4-videos submitted alongside this document. Each video shows the predictions of the proposed tracker TaMOs-SwinBase on the proposed LaGOT benchmark. Please note that we always produce a bounding box for visualization independent of its confidence score.

\begin{figure}[t]
    \centering
    \includegraphics[width=\linewidth, keepaspectratio]{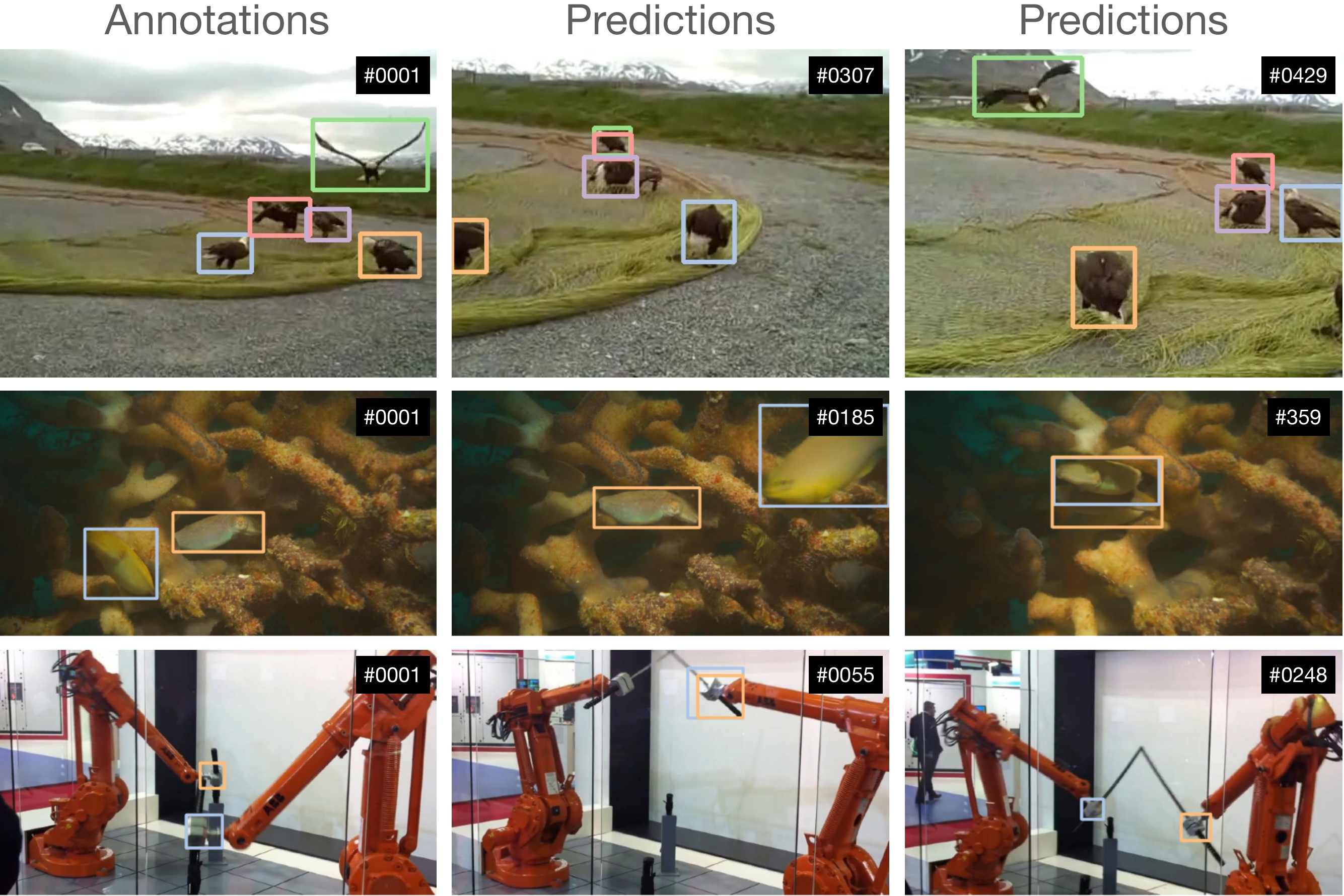}
\caption{Visual examples of failure cases of the proposed tracker (Ours-SwinBase) on different LaGOT sequences.}\label{sup:fig:failure_cases}
\end{figure}
\begin{table*}[t]
    \centering
    \newcommand{\best}[1]{\textbf{\textcolor{red}{#1}}}
    \newcommand{\scnd}[1]{\textbf{\textcolor{blue}{#1}}}
    \newcommand{\dist}{\hspace{10pt}}%
    \newcommand{\yes}{\textcolor{black}{\checkmark}}
    \newcommand{\no}{\textcolor{black}{\ding{55}}}
    \caption{Comparison of LaGOT and the existing datasets. Statistics is provided for test or validation set for the datasets for which test set annotations are hidden. \textsuperscript{*} For MOT15-20 we report stats on the train set.
    }\vspace{-0mm}
    \label{sup:tab:dataset-stats}%
    \resizebox{0.99\linewidth}{!}{%
        \begin{tabular}{l@{\dist}c@{\dist}c@{\dist}c@{\dist}c@{\dist}c@{\dist}c@{\dist}c@{\dist}c@{\dist}c@{\dist}}
        \toprule
                  & Num     & Num    & Avg Video length &  Avg Tracks  & Avg Track Length & Avg Track  & Avg Instances & Video & Annotation  \\
         Dataset  & Classes & Videos & (num frames)     &  per Video   & (num boxes)      & Length (s) & per frame     & FPS   & FPS  \\
        \midrule
        YouTubeVOS~\cite{Xu_2018_arXiv_youtubevos}                   & 91  & 474 & 135  & 1.74  & 27   & 4.5 & 1.64  & 30 FPS     & 6 FPS \\
        Davis17~\cite{DAVIS2017}                                     & -   & 30  & 67   & 1.97  & 67   & 2.8 & 1.97  & 24 FPS     & 24 FPS\\ 
        \midrule
        ImageNetVID\textsuperscript{*}~\cite{Jia_2009_CVPR_ImageNet} & 30  & 555 & 317  & 2.35  & 208 & 7    & 1.58  & 30 FPS     & 30 FPS\\
        \midrule
        TAO\textsuperscript{*}~\cite{Achal_2020_ECCV_TAO}            & 302 & 988 & 1010 & 5.55  & 21   & 21  & 3.31  & 30 FPS     & 1 FPS \\ 
        BDD100k~\cite{Yu_2020_CVPR_BDD100k}                          & 11  & 200 & 198  & 94.21 & 26   & 5   & 11.8  & 30 FPS     & 5 FPS \\ 
        MOT15~\cite{Dendorfer_2020_IJCV_MOTChallenge}\textsuperscript{*}                 & 1   & 11  & 500  & 45.5  &  75    & 3 & 8  & 2.5-30 FPS & 2.5-30 FPS\\
        MOT16~\cite{Dendorfer_2020_IJCV_MOTChallenge}\textsuperscript{*}                  & 1   & 7   & 760  & 74   &  273    &  10    & 38  & 14-30 FPS  & 14-30 FPS\\
        MOT20~\cite{Dendorfer_2020_IJCV_MOTChallenge}\textsuperscript{*}                 & 1   & 4   & 2233 & 583   &   572   &  23    & 150  & 25 FPS     & 25 FPS\\
        DogThruGlasses~\cite{huang_etal_cvpr23} & 1 & 30 & 419 & 3.3 & 352.6 & 11.7 & 2.4 & 30 FPS & 30 FPS \\
        \midrule
        GMOT-40~\cite{Bai_2021_CVPR_GMOT}                            & 10  & 40  & 240  & 50.65 & 133  & 5.3 & 26.6  & 24-30 FPS  & 24-30 FPS \\ 
        \midrule
        TrackingNet~\cite{2018_Muller_Trackingnet}                   & 27  & 511 & 442  & 1     & 442  & 15  & 1     & 30 FPS     & 30 FPS\\
        UAV123~\cite{Mueller_2016_ECCV_UAV123}                       & 8   & 123 & 915  & 1     & 915  & 28  & 1     & 30 FPS     & 30 FPS\\
        OTB-100~\cite{WU_2015_TPAMI_OTB}                             & 16  & 100 & 590  & 1     & 590  & 20  & 1     & 30 FPS     & 30 FPS \\
        NFS-30~\cite{Galoogahi_2017_ICCV_NFS}                        & 15  & 100 & 479  & 1     & 479  & 14  & 1     & 30 FPS     & 30 FPS\\
        GOT10k~\cite{Huang_2021_TPAMI_GOT10k}                        & 84  & 420 & 150  & 1     & 150  & 15  & 1     & 10 FPS     & 10 FPS\\
        % OxUvA~\cite{Valmadre_2018_ECCV_OxUvA}                      & 8   & 185 & 4198 & 1.08  & 60   & 140 & 1.08  & 30 FPS / 1 FPS \\ 
        OxUvA~\cite{Valmadre_2018_ECCV_OxUvA}                        & 8   & 200 & 4198 & 1     & 60   & 140 & 1     & 30 FPS     & 1 FPS \\
        LaSOT~\cite{Fan_2020_IJCV_Lasot_ext}                         & 71  & 280 & 2430 & 1     & 2430 & 81  & 1     & 30 FPS     & 30 FPS\\
        \midrule
        \textbf{LaGOT}                                               & 102 & 294 & 2258 & 2.89  & 707  & 71  & 2.41 & 30 FPS      & 10 FPS \\
        \bottomrule
        \end{tabular}
	}%
\end{table*}
\begin{figure}[t]
    \centering
    \includegraphics[width=\linewidth, keepaspectratio]{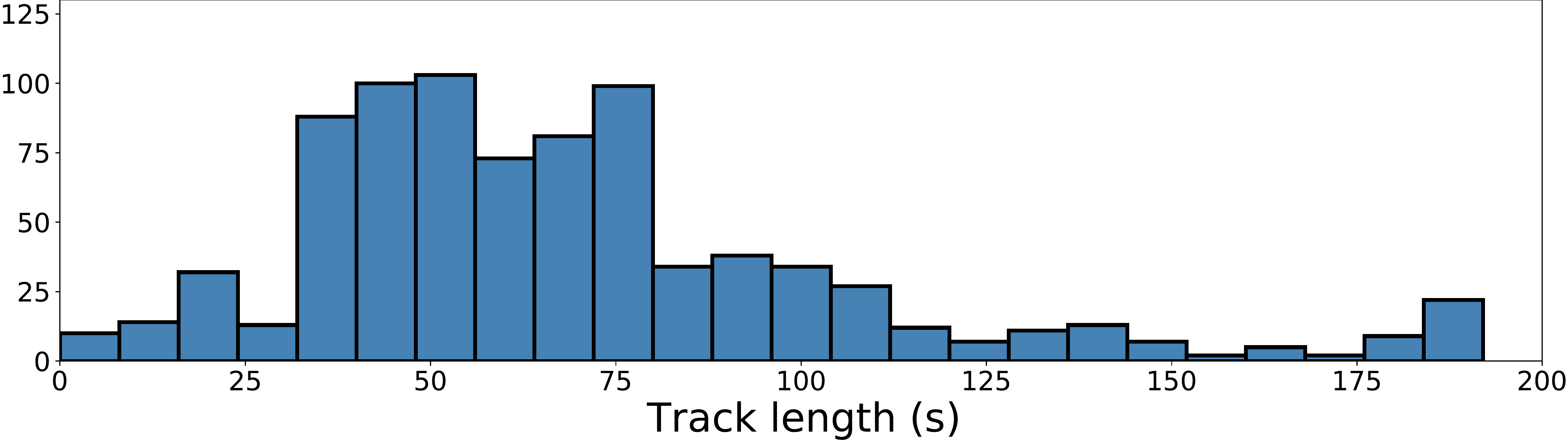}
\caption{Track lengths distribution of the LaGOT benchmark.}\label{sup:fig:track_length_dist}
\end{figure}
\begin{figure}[t]
    \centering
    \includegraphics[width=\linewidth, keepaspectratio]{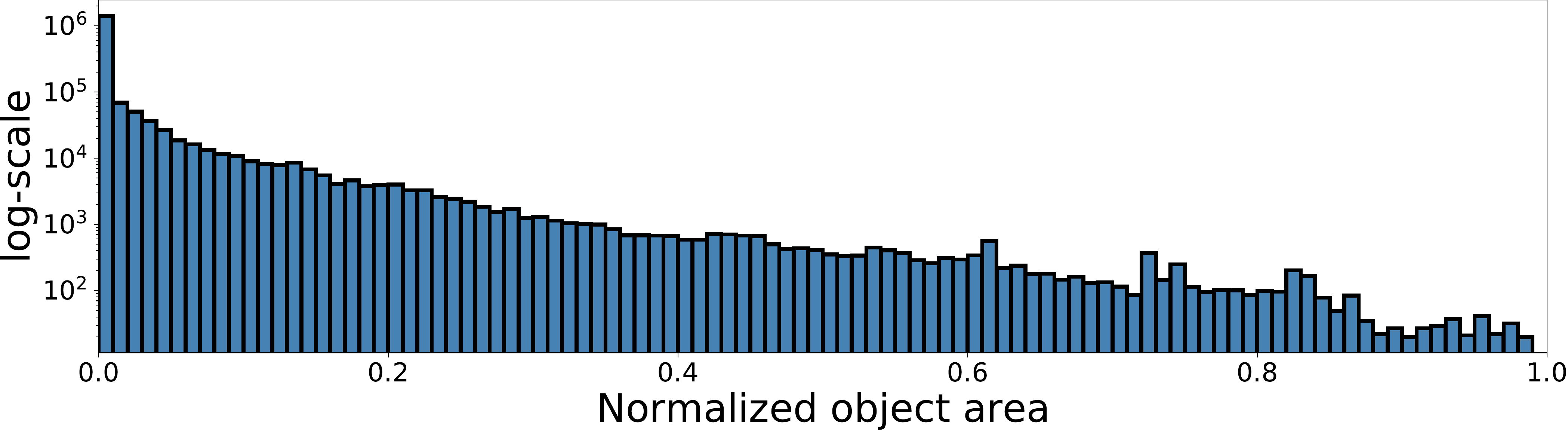}
\caption{Object size distribution of the LaGOT benchmark.}\label{sup:fig:object_sizes}
\end{figure}

\parsection{Failure Cases} Fig.~\ref{sup:fig:failure_cases} shows typical failure cases of the proposed tracker on three different sequences of the proposed LaGOT benchmark. Particularly challenging are videos that contain multiple visually similar objects since our tracker does not employ any motion model but rather tracks the objects via the learned appearance from the first frame. Another failure case occurs when the target object is no longer visible such that our tracker might start to track a visually similar distractor instead. However, once the target reappears our globally operating tracker is usually able to re-detect it. Lastly, if multiple visually similar objects need to be tracked our tracker might fail to distinguish these objects such that it produces multiple bounding boxes with different ids for the same object.

\section{Limitations and Future Work}\label{sup:sec:limitations}
Currently the number of objects that can be tracked is limited by the pool-size of the object embeddings. While it is possible to learn a larger pool-size it is cumbersome. Thus, an interesting direction for future research would be to generate an arbitrary number of object embedding on the fly such that any number of target objects can be tracked. 

Furthermore, we propose to use an \gls{fpn} to regress more accurate bounding boxes for small objects and show that adding such an \gls{fpn} helps. However, as in object detection, tracking extremely small objects is challenging due to the limited feature resolution when processing the full frame. 

\section{Datasets}\label{sup:sec:datasets}

Below we provide additional details about our annotated dataset, such as examples of new classes and various statistics, as well as an extensive comparison to existing datasets that focus on related tasks.

\subsection{Insights}
Fig.~\ref{sup:fig:track_length_dist} shows the distribution of the track lengths in seconds for all tracks in the proposed benchmark LaGOT. We observe that most tracks are between 30 and 110 seconds long. Furthermore, Fig.~\ref{sup:fig:object_sizes} shows the size distribution of the annotated objects in the dataset. We conclude that various sizes are present in the dataset but large objects are rare than small ones. Further, the distribution shows that the targets are not visible in a large amount of video frames indicated by an object area of zero.

During the annotation process, we added $31$ new classes: \textit{rotor}, \textit{fish}, \textit{backpack}, \textit{motor}, \textit{wheel}, \textit{garbage}, \textit{drum}, \textit{accordion}, \textit{super-mario}, \textit{hockey puck}, \textit{hockey stick}, \textit{kite-tail}, \textit{ball}, \textit{crown}, \textit{stick}, \textit{spiderweb}, \textit{head}, \textit{banner},\textit{face}, \textit{bench}, \textit{tissue-bag}, \textit{para glider}, \textit{star-patch}, \textit{shadow}, \textit{bucket}, \textit{helicopter}, \textit{sonic}, \textit{hero}, \textit{ninja-turtle}, \textit{reflection}, \textit{rider}.

\subsection{Comparison}

We provide a detailed comparison of related existing datasets in Tab.~\ref{sup:tab:dataset-stats}. We divide the table into Video Object Segmentation~(VOS), Video Object Detection, Multiple Object Tracking~(MOT), Generic Multiple Object Tracking~(GMOT) and Single Object Tracking~(SOT) datasets.

The length of VOS sequences is much shorter than in our LaGOT benchmark (2.8s/4.5s vs 71s). Similarly the video object detection dataset ImagenetVID contains shorter sequences (7s vs. 71s), fewer classes (30 vs 102) and a smaller number of average tracks per sequence (2.35 vs 2.89) than LaGOT. 
MOT datasets typically focus on fewer classes, contain shorter sequences or are annotated at low frame rates only.
TAO contains many more classes than typical MOT datasets but provides annotations only at 1 FPS leading to a much lower average number of annotated frames per track than LaGOT (21 vs. 707).
The GMOT-40 dataset contains fewer classes, fewer videos, shorter sequences and provides due to its task only annotations of one particular object class per sequence compared to LaGOT. 
In contrast to SOT datasets that provide only a single annotated object per sequence, LaGOT provides on average 2.89 tracks per sequence. Furthermore, it contains longer sequences than most listed SOT datasets.
Overall LaGOT enables to properly evaluate the robustness and accuracy of multiple object GOT methods. A key factor are the multiple annotated tracks per sequence at a high frame rate and the relatively long sequences.

\end{appendices}

\end{document}